%% file: acl_latex.tex
\documentclass[11pt]{article}

\usepackage{acl}
\usepackage{booktabs}
\usepackage{times}
\usepackage{latexsym}
\usepackage{makecell} 
\usepackage{placeins}

\usepackage[T1]{fontenc}
\usepackage[utf8]{inputenc}
\usepackage{tabularx}
\usepackage{microtype}
\usepackage{amsmath}
\usepackage{amssymb}
\usepackage{bbm}
\usepackage{inconsolata}
\usepackage{pifont}
\newcommand{\cmark}{\textcolor{green}{\ding{51}}} % green check mark
\newcommand{\xmark}{\textcolor{red}{\ding{55}}} % red cross mark
\usepackage{graphicx}
\usepackage[dvipsnames, svgnames, x11names]{xcolor}
\usepackage{multirow}
\usepackage{colortbl}

\definecolor{lightgray}{gray}{0.95}
\definecolor{deepblue}{RGB}{70,130,180}
\definecolor{deepgray}{RGB}{119,136,153}
\definecolor{RosyBrown}{RGB}{188,143,143}
\definecolor{PeachPuff3}{RGB}{205,175,149}

\usepackage[utf8]{inputenc}
\usepackage{xcolor}
\usepackage{listings}
\usepackage[most]{tcolorbox}
\usepackage{caption}
\lstdefinestyle{prompt}{
    basicstyle=\ttfamily\fontsize{7pt}{8pt}\selectfont,
    frame=none,
    breaklines=true,
    backgroundcolor=\color{lightgray},
    breakatwhitespace=true,
    breakindent=0pt,
    escapeinside={(*@}{@*)},
    numbers=none,
    numbersep=5pt,
    xleftmargin=5pt,
    aboveskip=2pt,
    belowskip=2pt,
}

\tcbset{
  aibox/.style={
    top=10pt,
    colback=white,
    enhanced,
    center,
  }
}
\newtcolorbox{AIbox}[2][]{aibox, title=#2,#1}

\usepackage[dvipsnames, svgnames, x11names]{xcolor}

\newcommand{\modelnamec}{%
    \textit{\textbf{%
    \textcolor[RGB]{0, 102, 204}{U}%     % 深蓝
    \textcolor[RGB]{51, 102, 255}{n}%    % 亮蓝
    \textcolor[RGB]{102, 0, 204}{i}%     % 靛紫
    \textcolor[RGB]{153, 51, 255}{C}%    % 丁香紫
    \textcolor[RGB]{204, 0, 153}{o}%     % 深玫红
    \textcolor[RGB]{255, 0, 127}{r}%     % 玫红
    \textcolor[RGB]{255, 51, 153}{n}%    % 亮粉玫
    }}%
}
\newcommand{\modelname}{\textbf{{UniCorn}}}

\newcommand{\mismatch}{\textbf{Conduction Aphasia}}
\newcommand{\evalname}{\textbf{{UniCycle}}}
\title{\raisebox{-4pt}{\includegraphics[scale=0.05]{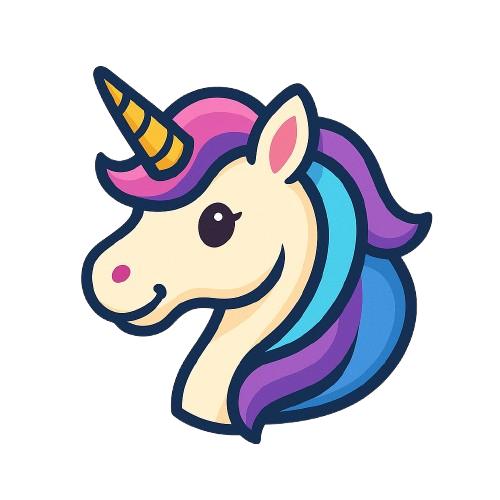}} \hspace{-1pt}\modelnamec: Towards Self-Improving Unified Multimodal Models through Self-Generated Supervision}

\usepackage{fontawesome5} % 提供 GitHub, 官网等图标
\usepackage{graphicx} % 用于插入自定义图标图片（如 Hugging Face）

\author{Ruiyan Han$^{2*}$ \quad Zhen Fang$^{1}\thanks{\quad Equal Contribution}\thanks{\quad Project Lead.}$ \quad  Xinyu Sun$^{2*}$ \quad Yuchen Ma$^{2}$ \quad Ziheng Wang$^{2}$ \quad Yu Zeng$^{1}\footnotemark[2]$ \\
\textbf{Zehui Chen$^{1}$ \quad  Lin Chen$^{1}$ \quad Wenxuan Huang$^{3,4}$ \quad Weijie Xu$^{5}$ \quad Yi Cao$^{6}$ \quad Feng Zhao$^{1}\thanks{\quad Corresponding author}$}\\
  $^{1}$MoE Key Lab of BIPC, USTC\quad $^{2}$FDU \\
$^{3}$ECNU\quad$^{4}$CUHK\quad$^{5}$NJU\quad$^{6}$SUDA \\
\vspace*{2cm} % 增加与上方的间距
\centering
\urlstyle{same} % 确保链接字体与正文字体一致
\href{https://github.com/Hungryyan1/UniCorn}{\faGithub \ \textbf{Code}} \quad \quad
\href{https://huggingface.co/CostaliyA/UniCorn}{\faRobot \ \textbf{Model}} \quad \quad
\href{https://github.com/shierlouz/Unicycle}{\faChartBar \ \textbf{Benchmark}} \quad \quad
\href{https://costaliya.github.io/UniCorn.github.io/}{\faGlobe \ \textbf{Project Page}}
\vspace{0.4cm}
}
\begin{document}
\maketitle
\input{sections/0_abs}
\input{sections/1_intro}

\input{sections/2_related}

\input{sections/3_method}

\input{sections/4_exp}

\input{sections/5_conclusion}

\section*{Limitations}
Despite achieving robust performance in both T2I generation and multimodal understanding, UniCorn possesses certain limitations. First, the current self-improvement framework operates in a single-turn manner and primarily enhances generative capabilities, with no significant gains observed in understanding metrics. In future work, we intend to explore multi-turn iterative self-play to foster the co-evolution of both capabilities. Second, the self-play mechanism requires the UMM to handle prompt generation, rollout, and judgment, which inevitably introduces additional computational costs. We plan to investigate more efficient methodologies to streamline this process in subsequent research.
\section*{Ethical Statement}
The development of UniCorn adheres to ethical standards for AI research. We utilize publicly available open-source models as our foundation and conduct all experiments using standard public benchmarks. Our self-improvement framework aims to enhance generative quality through internal feedback, thereby reducing the need for massive external data collection. While we implement internal filters during the self-play process to improve output alignment, we acknowledge that multimodal models may still reflect biases present in their pre-training data. We are committed to transparency and encourage the responsible use of our framework in downstream applications.
\bibliography{custom}
\input{sections/99_appendix}
% \appendix

% \section{Example Appendix}
% \label{sec:appendix}

% This is an appendix.

\end{document}

%% file: sections/0_abs.tex
\begin{abstract}
While Unified Multimodal Models (UMMs) have achieved remarkable success in cross-modal comprehension, a significant gap persists in their ability to leverage such internal knowledge for high-quality generation. We formalize this discrepancy as \textit{Conduction Aphasia}, a phenomenon where models accurately interpret multimodal inputs but struggle to translate that understanding into faithful and controllable synthesis. To address this, we propose {\modelnamec}, a simple yet elegant self-improvement framework that \textbf{ eliminates the need for external data or teacher supervision}. By partitioning a single UMM into three collaborative roles: Proposer, Solver, and Judge, {\modelname} generates high-quality interactions via self-play and employs cognitive pattern reconstruction to distill latent understanding into explicit generative signals. To validate the restoration of multimodal coherence, we introduce {\evalname}, a cycle-consistency benchmark based on a $Text \rightarrow Image \rightarrow Text$ reconstruction loop. Extensive experiments demonstrate that {\modelname} achieves comprehensive and substantial improvements over the base model across six general image generation benchmarks. Notably, it achieves \textbf{state-of-the-art (SOTA)} performance on TIIF(73.8), DPG(86.8), CompBench(88.5), and {\evalname}(46.5), while further delivering substantial gains of \textbf{+5.0} on WISE and \textbf{+6.5} on OneIG. These results highlight that our method significantly enhances T2I generation while maintaining robust comprehension, demonstrating the scalability of fully self-supervised refinement for unified multimodal intelligence.
\end{abstract}

%% file: sections/1_intro.tex
\section{Introduction}
The realization of Artificial General Intelligence (AGI) requires a tight synergy between comprehension and generation, wherein comprehension enables the internalization of knowledge and generation allows its coherent and expressive externalization.
By integrating multiple modalities into a shared representational space, Unified Multimodal Models (UMMs)~\cite{bagel,blip3o, xie2025show} naturally couple comprehension and generation as two complementary phases of a unified cognitive process, supporting both knowledge grounding and coherent reasoning.

\begin{figure}[!t]
    \centering
    \includegraphics[width=1.0\columnwidth]{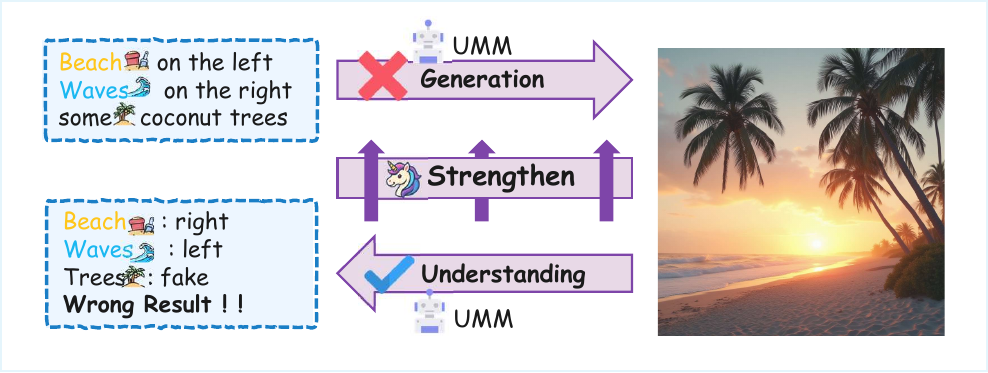}
      \vspace{-2em}
  \caption{\label{fig:intro}
\textbf{Motivation of {\modelname}}. UMMs often exhibit an understanding-generation gap: they can accurately understand and critique errors in an image, yet fail to generate the same scene correctly. This conduction aphasia motivates our framework to leverage the model’s superior internal understanding to strengthen and refine its generative capabilities through self-contained feedback.}
     % \vspace{-1.7em}
          \vspace{-1em}
\end{figure}

Despite these advances, a fundamental disparity remains between comprehension and generation in current UMMs. This mismatch, which we formalize as {\mismatch}, arises when a model demonstrates strong domain understanding yet fails to translate that knowledge into high-quality generative outputs. As shown in Fig.~\ref{fig:intro}, a representative case appears in image generation: although the model can accurately recognize what an image depicts and reliably assess its visual quality, it often cannot act on this knowledge during generation. This disconnect motivates a central research question: \textbf{\textit{how can a model’s robust understanding guide and strengthen its generative behavior?}}

Driven by this simple yet fundamental question, we propose {\modelnamec}, a post-training framework that enables self-improvement through a unified cycle of proposal, execution, and evaluation. Requiring no external data or teacher-model supervision, {\modelname} allows UMMs to autonomously narrow the comprehension--generation gap by acting as their own instructor within a single parameter space. Motivated by the observation that a single UMM can exhibit distinct capabilities for proposing, executing, and evaluating, we treat the model as a modular system in which comprehension can explicitly guide generation. This design turns the model’s latent interpretive capability into an internal training signal, enabling autonomous generative improvement without external supervision.

Specifically, {\modelname} operates through a self multi-agent framework that functionalizes the UMM into three distinct internal roles. The process begins with the model acting as a \textbf{Proposer} to propose diverse and expansive prompts, followed by its transition into a \textbf{Solver} to synthesize corresponding image candidates. Finally, it assumes the role of a \textbf{Judge} to provide evaluative rewards based on its superior comprehension.

By simulating structured collaboration within a single parameter space, this design yields rich interaction data that are refined through data reconstruction. Concretely, we convert raw multi-agent outputs into structured training signals, including descriptive captions, evaluative judgments, and reflective feedback, thereby distilling latent understanding into explicit supervision for effective self-improvement.

To determine whether internal collaboration produces general multimodal intelligence instead of narrow task fitting, we introduce {\evalname}, a cycle-consistency benchmark that probes cognitive alignment via informational integrity. Existing evaluations often separate comprehension and generation, which can lead to piecemeal measurements and biased conclusions. In contrast, {\evalname} frames evaluation as a Text $\rightarrow$ Image $\rightarrow$ Text reconstruction process. It compares the model’s original intent with its reconstructed description, using the resulting semantic gap as a holistic, training-free indicator of conceptual coherence, while reducing the bias that arises when capabilities are tested in isolation.

Across extensive experiments, we find that our model achieves reliable self-improvement without heuristic reward engineering, curriculum design, or external supervision. Compared with prior self-improvement approaches~\cite{srum} and methods that depend on external guidance, our approach learns from internally generated training signals, generalizes well, and remains stable under out-of-distribution (OOD) conditions. These results support the effectiveness of a fully self-contained learning paradigm.
\begin{figure}[t]
    \centering
\includegraphics[width=0.4\textwidth]{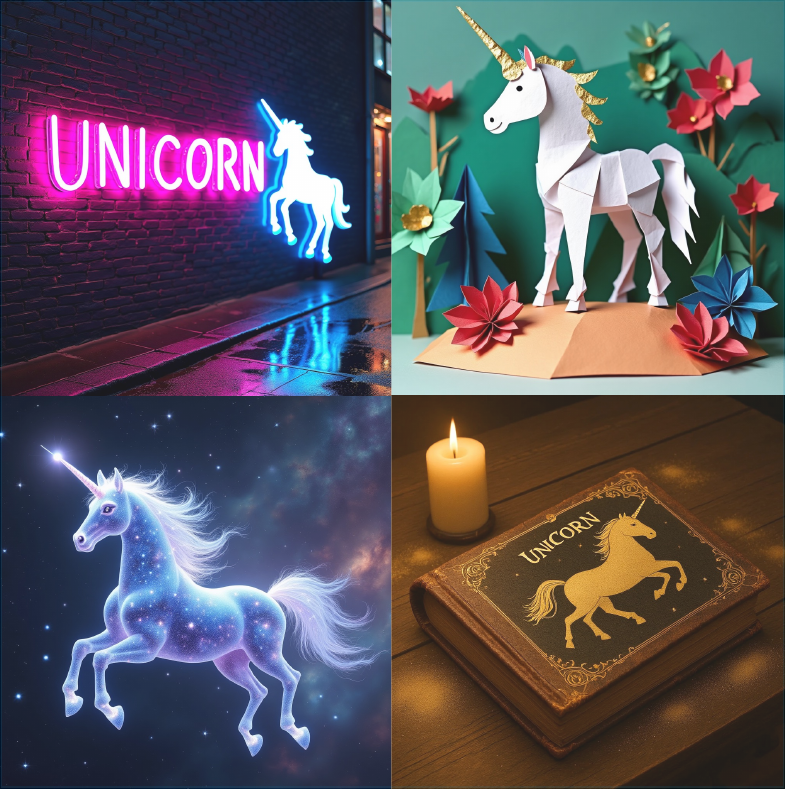}
    \caption{Visualization results of {\modelname}.}
    \label{fig: unicorns}
\end{figure}

\begin{itemize}
\item We identify the {\mismatch} phenomenon in UMMs, where strong understanding fails to translate into accurate generation, and propose {\modelnamec}, which repurposes internal comprehension as self-supervision through Proposer, Solver, and Judge roles with data reconstruction.
\item To assess whether multimodal understanding and generation remain conceptually consistent across modality transitions, we introduce {\evalname}, a training-free evaluation protocol that measures multimodal coherence through a Text $\rightarrow$ Image $\rightarrow$ Text cycle.
\item Experimental results demonstrate that our method consistently outperforms prior approaches, achieving \textbf{SOTA} performance on TIIF (73.8), DPG (86.8), CompBench (88.5), and {\evalname}(46.5), together with substantial improvements of \textbf{+4.0} on Geneval, \textbf{+5.0} on WISE, and \textbf{+6.5} on OneIG.

\end{itemize}

%% file: sections/2_related.tex
\section{Related Work}
\paragraph{Unified Multimodal Models}
UMMs aim to unify cross-modal understanding and generation, yet strong understanding often fails to yield equally strong native generation. Existing designs fall into two paradigms: \emph{pure autoregressive} models that jointly predict text and visual tokens over interleaved sequences ~\cite{chen2025janus,cui2025emu3,tong2025metamorph}) and \emph{hybrid} models that combine autoregressive language modeling with diffusion-based image synthesis, either within a unified backbone ~\cite{xie2024show,zhao2024monoformer}) or via modular routing and sparse experts~\cite{shi2024lmfusion,liang2024mixture,deng2025emerging}), with related guidance schemes such as Diffusion Forcing~\cite{chen2024diffusion}. Beyond architecture, self-improvement methods convert self-generated signals into training objectives~\cite{yu2025guided,zhou2024calibrated,wang2025unified}; for UMMs, SRUM derives internal rewards from understanding~\cite{srum}, and UniRL jointly optimizes understanding and generation~\cite{unirl}. However, most pipelines depend on auxiliary components or task-specific feedback, limiting scalability and generalization.

\paragraph{Multi-Agent and Self-Improvement Learning}
Multi-agent systems decompose reasoning through role specialization and interaction, enabling solution diversity and cross-verification, but often incur high coordination cost and brittle verification~\cite{chen2024agentverse,liang2024encouraging,cemri2025multi}. In parallel, LLM self-improvement converts self-generated tasks and evaluations into training signals, supporting zero-data learning via self-play and self-rewarding mechanisms~\cite{silver2017mastering,huang2025r,zhao2025absolute,yuan2024self}. Unified Multimodal Models (UMMs) naturally unify understanding and generation within a single parameter space, making them particularly well-suited for lightweight role instantiation and fully model-driven self-improvement without external supervision.

%% file: sections/3_method.tex
\section{Method}
In this section, we begin by presenting the motivation through an analysis of the mismatch between generation and understanding capabilities in UMMs. Building on these observations, we introduce {\modelnamec}, a simple yet elegant post-training framework that enables self-improvement without any external annotated data or teacher models.

\subsection{Motivation}
Just as a child who associates the word “apple” with the fruit can spontaneously name it upon seeing it, cognitive symmetry~\cite{blanco2018unconscious} enables a bidirectional mapping between internal concepts and external expressions. This alignment is reminiscent of escaping Plato’s Cave: true intelligence must move beyond observing surface data to mastering the reciprocal relationship between an appearance and its underlying source.

However, current UMMs suffer from a functional deficit akin to {\mismatch}: while the model exhibits profound comprehension, its generative performance remains fractured, failing to produce the very content it can inherently understand. Bridging this gap is critical; without aligning these dual processes, a model remains a "passive observer," capable of grounding symbols but incapable of utilizing them. Mastering the synergy between understanding and generation is thus not merely a functional upgrade but the essential step toward achieving the cognitive integrity required for AGI.

\begin{figure}[!t]
    \centering
    \includegraphics[width=1.0\columnwidth]{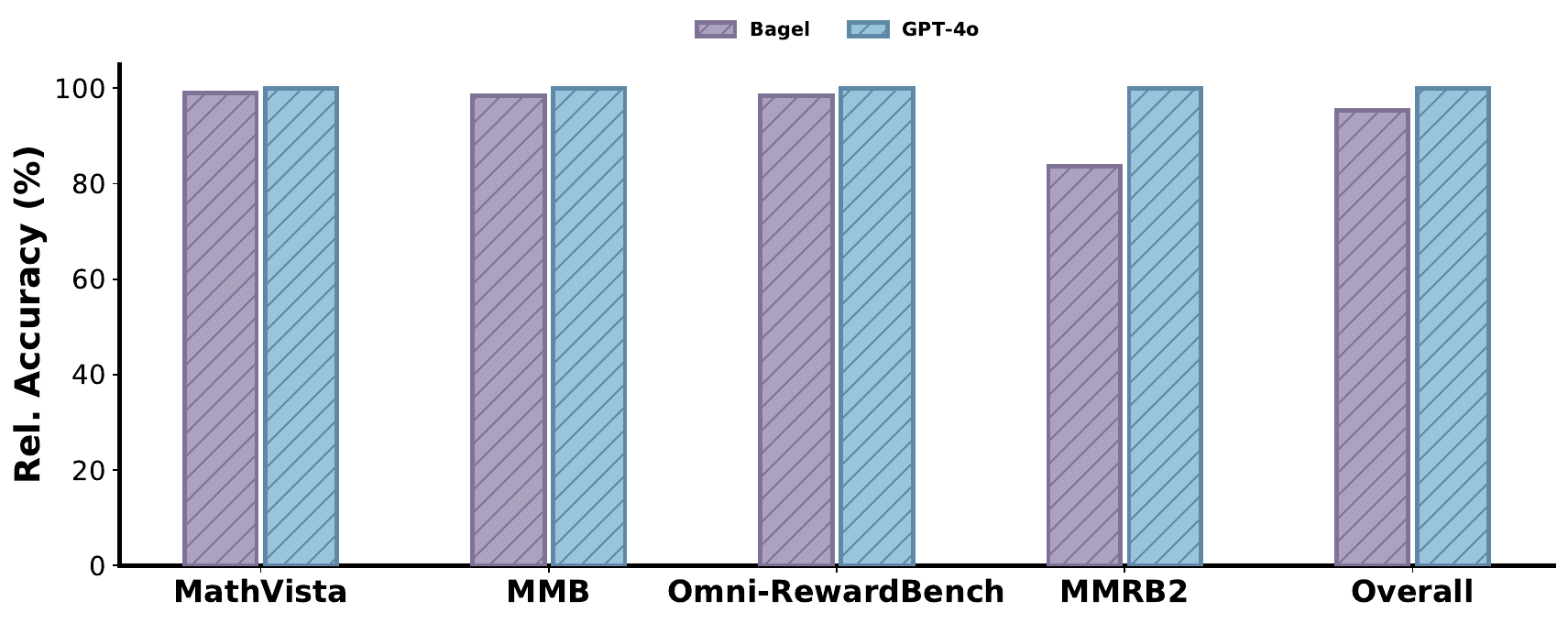}
        \vspace{-2em}
      \caption{\label{fig:motivation} \textbf{Results of BAGEL~\cite{bagel} and GPT-4o~\cite{gpt4o} on four understanding benchmarks.} For Omini-RewardBench~\cite{jin2025omni} and MMRB2~\cite{hu2025multimodal}, we evaluate the T2I task.  Performances are normalized with GPT-4~\cite{achiam2023gpt}
results for better visualization.}
    \vspace{-1.5em}
\end{figure}

On the one hand, as illustrated in Fig.~\ref{fig:motivation}, current UMMs demonstrate formidable perception and comprehension capabilities. Specifically, when serving as a reward model for Text-to-Image (T2I) generation, the UMM exhibits a sophisticated grasp of cross-modal semantics. This suggests that the model has already internalized a robust 'world model' and possesses the necessary latent knowledge to discern high-quality visual-textual alignments.

On the other hand, the model’s generative capability remains markedly constrained, primarily due to its failure to bridge the gap between internal recognition and active synthesis. This functional dissociation means that the UMM’s own sophisticated understanding remains a 'silent passenger' during the generative process, unable to inform or correct its outputs. Building on this observation, our key insight is that \textbf{the UMM’s formidable comprehension can be repurposed as an autonomous supervisory signal to steer its generative behavior}. By transforming latent interpretive depth into explicit guidance, we promote a tighter coupling between these two processes, ultimately restoring the cognitive symmetry essential for a truly integrated multimodal intelligence.

\subsection{Problem Definition}
\begin{figure*}[!t]
    \centering
    \includegraphics[width=\textwidth]{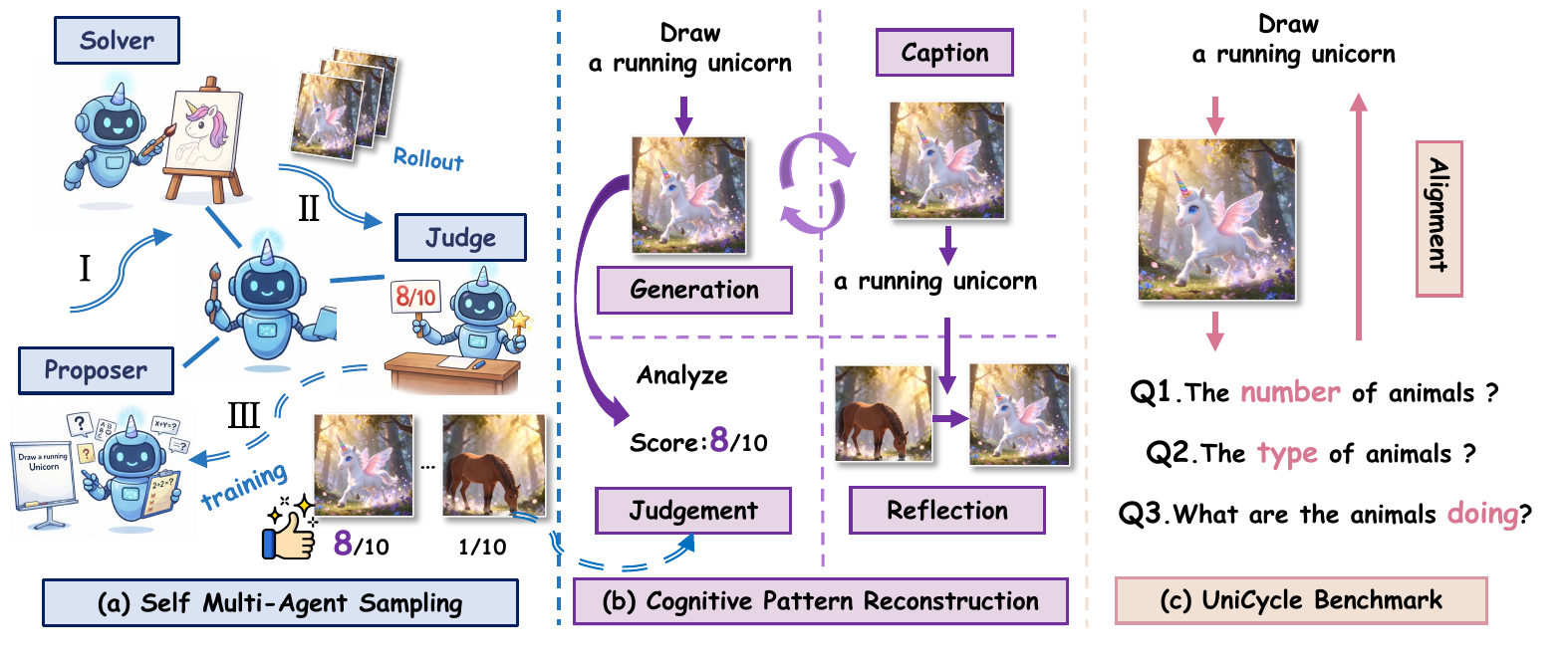}
        \vspace{-2em}
      \caption{\label{fig:framework} \textbf{Overview of the UniCorn Framework}. (a) Illustrates the self-multi-agent collaboration for high-quality data sampling. (b) Details the Cognitive Pattern Reconstruction process, which reorganizes data to facilitate robust and efficient learning. (c) Presents the UniCycle benchmark evaluation, verifying whether the model can accurately reconstruct key textual information from its own generated content.}
    \vspace{-1em}
\end{figure*}
We study UMMs that process interleaved image-text inputs and outputs. 
A UMM is formulated as a policy $\pi_\theta$ that maps a multimodal input sequence
\begin{equation}
X = (x_1,\ldots,x_N), \quad x_n \in T \cup I,
\end{equation}
to an interleaved multimodal output sequence $Y = \pi_\theta(X)$.
This unified input-output formulation supports both  Image-to-Text (I2T) understanding and  Text-to-Image (T2I) generation.
We operationalize understanding as I2T and generation as T2I, and leverage the model’s stronger I2T understanding to supervise and refine its weaker T2I generation.

\subsection{{\modelname}}
{\modelnamec} operates via two core stages: Self Multi-Agent Sampling and Cognitive Pattern Reconstruction (CPR). First, the UMM concurrently assumes three roles: Proposer, Solver, and Judge (\S~\ref{sec:roles}), to simulate a collaborative loop. Then, the CPR stage reconstructs these raw interactions into three training patterns: caption, judgement, and reflection (\S~\ref{sec:re-data}), which are combined with high-quality self-sampled T2I generation data for post-training. Critically, the entire process is \textbf{fully self-contained, requiring no external teacher models or human-annotated data}.

\subsubsection{Stage 1: Self Multi-Agent Sampling}
\label{sec:roles}
LLMs are naturally suited for self-play in multitask settings~\cite{radford2019language}. For UMMs, interleaved multimodal inputs and functional diversity allow prompting, generation, and judgement to coexist within a shared model, enabling role-conditioned behaviors under different prompts. We leverage this property to functionalize a single UMM into collaborative roles, bridging the comprehension--generation gap through internal synergy.
\paragraph{Proposer $\pi_\theta(T \mid T)$}
The proposer is designed to generate a diverse set of challenging prompts for the unified multimodal model, which are subsequently used to produce training images.
To this end, inspired by LAION-5B~\cite{laion-5b} and COYO-700M~\cite{coyo-700m}, we partition all T2I task prompts into ten categories and designed fine-grained generation rules for each category.
Next, we prompt UMM to generate an initial batch of prompts and act as the judge to select the best candidate for subsequent iterations.
Leveraging the strong in-context learning (ICL) capabilities of LLMs~\cite{dong2024survey}, the initial example serves as a few-shot demonstration to guide the generation of subsequent prompts. To further enhance diversity, we introduce a dynamic seeding mechanism. After generating a predefined number of prompts, several examples are sampled from the prompt library for evaluation and then used to construct new demonstrations that guide the next round of prompt generation.
Compared with prior approaches that either directly rely on training set~\cite{srum} or employ external models for prompt construction~\cite{unirl}, our method requires no external data and generates more diverse prompts, thereby improving generalization.
\paragraph{Solver  $\pi_\theta(I \mid T)$}
The solver is responsible for producing a diverse set of outputs in response to the prompts generated by the proposer.
Therefore, we encourage the UMM to generate images under random seeds and different hyperparameters.
Following DeepSeek-R1~\cite{deepseek-r1}, we perform 8 rollouts per prompt to strike a favorable trade-off between sample quality, diversity, and computational efficiency.

\paragraph{Judge $\pi_\theta(T \mid T,I)$}
The judge is responsible for assigning scores to the images generated by the solver in response to prompts proposed by the proposer, which are then used for rejection sampling during training.

Previous work has relied on heuristic reward functions based on keywords~\cite{unirl} or on powerful external models to provide dense reward maps~\cite{srum}. Such reward judges depend heavily on parameter tuning and the performance of external models, which varies across tasks, thereby severely limiting the generalization of self-improvement.
As illustrated in Fig.~\ref{fig:motivation}, UMMs exhibit strong reward modeling capabilities. Thus, we formulate reward evaluation for all T2I tasks using discrete scores ranging from 0 to 10, following a widely adopted LLM-as-a-judge paradigm~\cite{radford2019language,kim2023prometheus}. To further enhance judgement quality, we transfer generation reward models~\cite{deepseek-grm}, which have demonstrated strong potential in LLMs, to T2I evaluation. Specifically, we design task-specific rubrics for each category and encourage the model to explicitly articulate its reasoning before producing the final score.

\subsubsection{Stage2: Cognitive Pattern Reconstruction}
\label{sec:re-data}
Through self multi-agent rejection sampling using the Proposer--Solver--Judge pipeline, we obtain a batch of high-quality prompt--image pairs.
While these pairs reflect a mapping from abstract conceptual spaces to high-dimensional visual manifolds, directly optimizing this cross-domain alignment remains stochastic and inefficient, often leading to mode collapse~\cite{chen2025t2i,wang2024div}. To move beyond this "black-box" optimization, we draw inspiration from metacognitive theory~\cite{dunlosky2008metacognition}, which identifies monitoring, evaluation, and regulation as the pillars of robust learning.
Based on this insight, we propose a tripartite data architecture that reclaims and structures the overlooked trajectories from the self-play cycle. By replaying these latent interactions as explicit caption, judgement, and reflection patterns, we respectively ground abstract concepts in visual features, provide evaluative signals, and encode self-correction processes. This design transforms the previously discarded internal "inner monologue" into a structured supervisory signal, fostering cognitive symmetry without external intervention.

\textbf{\textsc{Caption}} 
    To establish robust \textbf{semantic grounding}, this pattern ensures the model internalizes the conceptual essence of its own creations by optimizing the inverse mapping $\pi_\theta(T \mid I^*)$. By treating the highest-scoring image $I^*$ as the input and its originating prompt $T$ as the ground truth, the model learns to \textbf{anchor} abstract concepts within the specific visual manifolds it is capable of synthesizing, thereby reinforcing the bidirectional cognitive symmetry between internal concepts and external manifestations.

\textbf{\textsc{Judgement}} 
    This pattern focuses on \textbf{evaluative calibration} to refine the model's internal value system. We train the model to predict the evaluative signal $J$ for any generated pair, formulated as $\pi_\theta(J \mid T, I)$. By leveraging the task-specific rubrics and reasoning traces provided by the Judge, the model develops an acute perception of the latent gap between its current output and the ideal objective, providing a critical diagnostic signal for stabilizing the generative process.

\textbf{\textsc{Reflection}} 
    Inspired by Reflexion~\cite{shinn2023reflexion}, this pattern introduces \textbf{iterative regulation} to enhance the model's capacity for self-evolution. Leveraging the Solver's multiple rollouts $\{I_1, \dots, I_n\}$, we utilize the rewards assigned by the Judge to identify pairs of contrasting quality, specifically selecting a high-reward "winning" image $I^*$ and a lower-reward "losing" image $I_{lose}$ from the same prompt. We then construct reflection trajectories formulated as $\pi_\theta(I^* \mid T, I_{lose}, J)$, which explicitly encode the transition from suboptimal states to superior ones. By learning to transform the lower-quality manifestation $I_{lose}$ into its optimized counterpart $I^*$, the model internalizes a mechanism for self-correcting generative errors, effectively mitigating mode collapse without the need for external supervision.

These three data types are combined with high-quality self-sampled T2I generation data to fine-tune the UMM. Note that the whole reconstruction procedure is rule-based and does not introduce any complexity. Detailed generation pipeline and examples can be found in Appendix \ref{appd:training_detail}.

\subsection{{\evalname}}
\label{sec:unicycle}
To assess whether internal collaboration yields genuine multimodal intelligence rather than task-specific performance gains, we introduce \evalname{}, a cycle-consistency benchmark that measures information preservation under a \textbf{Text $\rightarrow$ Image $\rightarrow$ Text} loop. Given an instruction, \evalname{} evaluates whether a unified multimodal model can recover instruction-critical semantics from its \emph{own} generated image through subsequent visual understanding.
\input{tables/main_result}

Based on TIIF~\cite{wei2025tiifbenchdoest2imodel}, we generate QA pairs to probe instruction-implied attributes grounded in the generated image, extending the original TIIF benchmark from the T2I setting to the Text-to-Image-to-Text (T2I2T) setting. After annotation, we obtain 1,401 TIIF-style instances that cover more than ten task categories and span multiple question formats, including multiple-choice, binary (yes/no), and open-ended questions.

For evaluation, given a prompt $T$, the model first generates an image and then answers each question $q_k$ independently conditioned on the generated image.
An external judge model assesses whether each predicted answer $\hat{y}_k$ is consistent with the initial prompt $T$ and the reference answer $a_k$, and
produces a score for each question.

We define a unified metric to quantify this T2I2T consistency.

Let $\mathcal{Q}(T)$ denote the set of questions associated with a prompt $T$. We define
\begin{equation}
\begin{aligned}
\mathrm{Soft}(T)
&= \frac{1}{|\mathcal{Q}(T)|}\sum_{k \in \mathcal{Q}(T)} s_k, \\[-1pt]
\mathrm{Hard}(T)
&= \mathbbm{1}\!\left[\forall k \in \mathcal{Q}(T),\; s_k = 1\right].
\end{aligned}
\label{eq:cycle-soft-hard}
\end{equation}
 where $s_k$ denotes the judge score for question $q_k$, defined as a binary
indicator for non-text questions and as the proportion of correctly recovered
Keywords to enable a more fine-grained and continuous metric for text-type questions.

The final Soft and Hard scores are obtained by averaging over all prompts.
Additional details on data construction and evaluation prompt templates
are provided in Appendix~\ref{sec:unicycle details}.

%% file: tables/main_result.tex
\begin{table*}[t]
% \tiny
\scriptsize
% \footnotesize
\centering
\renewcommand{\arraystretch}{1.2}  % 缩小行高
\setlength{\tabcolsep}{3pt}        % 缩小列间距
\definecolor{mypurple}{HTML}{F5EDFC}
\definecolor{myred}{HTML}{FCEDF5}
% \definecolor{mygreen}{HTML}{F5FCED}
\definecolor{mygreen}{HTML}{EDFCED}
\begin{tabular}{lcc|ccc|ccc|ccc|c|c}
\Xhline{1.5pt}
\multirow{2}{*}{Model}
& \multicolumn{2}{c|}{\textbf{TIIF}  ↑}
& \multicolumn{3}{c|}{\textbf{WISE}  ↑}
& \multicolumn{3}{c|}{\textbf{OneIG-EN}  ↑}
& \multicolumn{3}{c|}{\textbf{CompBench}  ↑} 
& \multicolumn{1}{c|}{\textbf{DPG}  ↑}
& \multicolumn{1}{l}{\textbf{Geneval}  ↑}\\

& Short 
& Long 
& Physics
& Chemistry 
& \textbf{Overall}
& Text
& Alignment
& \textbf{Overall}
& Numeracy
& 3d Spatial
& \textbf{Overall} 
& Score
& Score\\
\hline

\rowcolor{AliceBlue}
\multicolumn{14}{l}{\textit{\textbf{Generation Only Models}}} \\

SD3 Medium
&64.8	 &64.8
& 47.0 & 29.0 & 42.0
& 40.7 &80.6  & 42.8
& 72.8 & 77.8 & 84.3 
& 84.1 & 74\\
FLUX.1 dev
& 66.2 &  66.7
&51.0  &35.0  &50.0 
& \underline{52.3} & 78.6 & 43.4
&75.3 &76.4  & 83.1 
& 83.8 & \underline{82}\\
\rowcolor{AliceBlue}
\multicolumn{14}{l}{\textit{\textbf{Unified Multimodal Models}}} \\
Janus-Pro
& 65.4 &61.1  
& 42.0 & 26.0 & 35.0
& 0.1 & 55.3 & 26.7
& 56.4 & 76.2 & 74.0 
& 84.3 & 80.0\\
show-o2
& 62.8  &63.9  
& \underline{63.0} & \textbf{49.0} & \textbf{61.0}
& 0.2 & \underline{81.7} & 30.8
& 69.7 & \textbf{88.6} & 82.8 
& \underline{86.1} & 76.0\\
BLIP3-o
& 58.8 & 58.7 
& \underline{63.0} & 37.0 & 52.0
& 1.3 & 71.1 & 30.7
& 71.7 & 81.7 & 84.7 
& 80.7 & \textbf{84.0} \\
OmniGen2
& 70.2 & 70.3  
& 52.0 & 34.0 & 48.0
& \textbf{68.0} & 80.4 & \textbf{47.5}
& 72.0 & 82.2 & \underline{85.8}
& 83.6 & 80.0\\

TwiG$^{\dagger}$
& - &   -
& - & - & -
& - & - & -
&61.9 & 38.9 & -
&- & -\\
T2I-R1
& 67.6 & 68.3  
& 55.0 & 30.0 & 54.0
&7.3  &80.4  &27.7 
& \underline{83.3} & 79.4 &81.9
& - &  77.0\\

\hline

BAGEL
& \underline{71.0} & \underline{71.8}  
& 57.0 & 43.0 & 50.0
& 24.4 & 76.9 & 36.1
& 70.4 & 78.0 & 82.2
& 84.0 & 78.0\\
\rowcolor{mygreen}
\modelname 
&\textbf{74.7}  & \textbf{72.9} &  \textbf{67.0} 	& \underline{47.0}  & \underline{55.0}
& 46.8& \textbf{84.1} & \underline{42.6}
&\textbf{83.5}&\underline{84.1}&\textbf{88.5}
& \textbf{86.8}& \underline{82.0}\\
\rowcolor{mygreen}
\textbf{$\triangle$(\textit{vs.} BAGEL)}
&\textbf{\textcolor{ForestGreen}{+3.7}}	  & \textbf{\textcolor{ForestGreen}{+1.1}}
& \textbf{ \textcolor{ForestGreen}{+10.0}}	& \textbf{\textcolor{ForestGreen}{+4.0}} & \textbf{\textcolor{ForestGreen}{+5.0}}
& \textbf{\textcolor{ForestGreen}{+22.4}} & \textbf{\textcolor{ForestGreen}{+7.2} }& \textbf{\textcolor{ForestGreen}{+6.5}}
&\textbf{\textcolor{ForestGreen}{+13.1}  }&\textbf{\textcolor{ForestGreen}{+6.1}}&\textbf{\textcolor{ForestGreen}{+6.3}}
&\textbf{\textcolor{ForestGreen}{+2.8}} &\textbf{\textcolor{ForestGreen}{+4.0}}\\

\bottomrule
\end{tabular}
\caption{\textbf{Evaluation results on TIIF, WISE, OneIG-EN, CompBench, DPG, and Geneval benchmarks}. Arrows (↑) denote that higher is better. \textbf{Bold} indicates the best performance across all models, and the second best is \underline{underlined}. The WISE score is normalized to a 0–100 scale for visualization. Detailed comparison is listed in Appendix~\ref{sec: detail_results}.}
\vspace{-4mm}
\label{tab: result_all}
\end{table*}

%% file: sections/4_exp.tex
\section{Experiments}
In this section, we first introduce the experiment setup, and conduct extensive experiments to demonstrate the effectiveness of our method.
\subsection{Experiment Setup}
\paragraph{Implementation }
We adopt BAGEL~\cite{bagel} as the base model for our main experiments.
The Proposer generates 5,000 prompts, then the Solver rolls out 8 times for each prompt.
Training is conducted for 600 steps on 8 NVIDIA H800 GPUs for about 7 hours with a constant learning rate of $1\times10^{-5}$.
Additional details are provided in the Appendix ~\ref{appd:training_detail}.

\begin{figure*}[h]
    \centering
    \includegraphics[width=\textwidth]{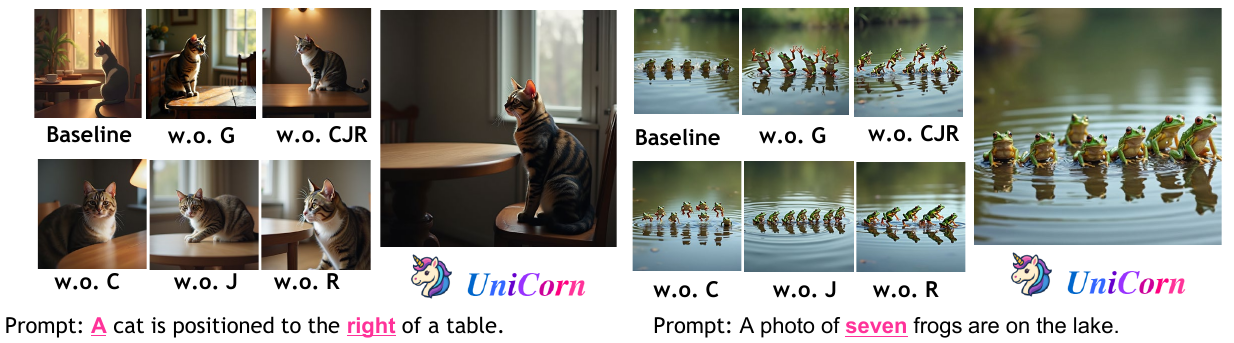}
    \caption{Qualitative comparison between \modelname, BAGEL  and \modelname's adifferent data settings. Our method jointly balances visual aesthetics, prompt fidelity, and realism in generation.}
    \label{fig: compare_model}
\end{figure*}
\paragraph{Baselines and Benchmarks}
To validate our method, we compare it against three categories of approaches. First, we consider baseline models, including T2I frameworks: SD3 Medium~\cite{SD3}, FLUX.1-dev~\cite{flux} and unified multimodal models: Janus-Pro~\cite{januspro2025}, Show-o2~\cite{xie2025show}, BLIP3-o~\cite{blip3o},  UniGen~\cite{tian2025unigen}, TwiG~\cite{guo2025thinking} and T2I-R1~\cite{jiang2025t2i}.
\begin{figure*}[!t]
    \centering
    \includegraphics[width=0.9\textwidth]{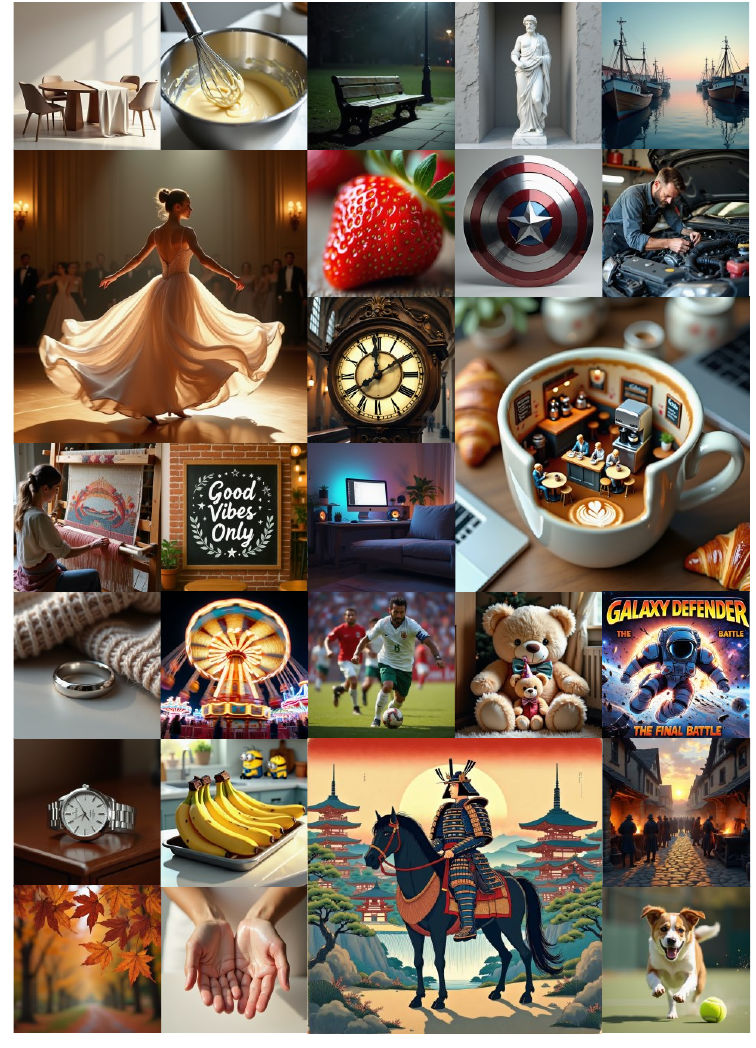}
      % \vspace{-2em}
  \caption{\label{fig: vis}
\textbf{Visualization results of {\modelname} at 1024×1024 resolution. }}
     % \vspace{-1.7em}
          % \vspace{-0.5em}
\end{figure*}
Regarding evaluation, we focus on TIIF~\cite{wei2025tiifbenchdoest2imodel}, WISE~\cite{niu2025wise}, OneIG-EN~\cite{chang2025oneigbenchomnidimensionalnuancedevaluation}, CompBench~\cite{kil2024compbench}, DPG~\cite{hu2024ella}, and Geneval~\cite{ghosh2023geneval} to assess generation performance. To evaluate understanding, we further report results on standard benchmarks including MME~\cite{fu2023mme}, MMB~\cite{liu2024mmbench}, MMMU-val~\cite{yue2024mmmu}, MMVP~\cite{tong2024eyes}, and MMStar~\cite{chen2024we}.

\subsection{Main Results}
As shown in Tab.~\ref{tab: result_all}, {\modelname} achieves highly competitive performance across five T2I benchmarks. Our method significantly enhances fine-grained instruction following on TIIF, particularly improving robustness to short prompts (+3.7 points). On the comprehensive OneIG benchmark, {\modelname} yields a 6.5-point overall improvement, with a remarkable 22.4-point gain in the Text subtask, indicating superior internalization of underlying knowledge. Furthermore, {\modelname} achieves a 5 point gain on the knowledge-intensive WISE benchmark and a 6.3 point boost on CompBench. Notably, the substantial improvements in Numeracy (+13.1) and 3D Spatial (+6.1) tasks demonstrate the effective transfer of structured understanding into faithful synthesis, with {\modelnamec} even surpassing \textbf{GPT-4o} on DPG benchmark (86.8 vs 86.2). These results consistently demonstrate that our self-play framework enables UMMs to bridge the gap between multimodal understanding and controllable generation, achieving robust performance that rivals state-of-the-art closed-source models.

\subsection{Ablation Study}

%scale up
This section conducts ablation studies on data pattern, model architecture, and dataset size to further analyze our method.
\subsubsection{Data Pattern}
\input{tables/data_ablation}

This section deconstructs multimodal data patterns to demonstrate how  Cognitive Pattern Reconstruction bridges the gap between understanding and generation within a unified framework.

Tab.~\ref{tab:ablation_gcjr} reveals a hierarchical synergy between data patterns: while relying solely on generation (w.o. CJR) maintains basic instruction following (TIIF-S: 72.3), it triggers a catastrophic collapse of the latent space, evidenced by the sharp drop in MME-P from 1685.0 to 311.0. This proves that unconstrained generative training without semantic grounding leads to mode collapse. Conversely, incorporating Cognitive Pattern Reconstruction patterns (C, J, R) stabilizes the model; Judgment and Reflection provide evaluative signals that boost complex generative quality (TIIF-R: 78.4), while Captioning preserves the multimodal foundation and spatial reasoning capabilities. Finally, although removing generation (w.o. G) maintains comprehension metrics like MME-P (1669.0), it stalls generative growth, resulting in lower TIIF scores (73.4). 
Qualitative comparisons are shown in Fig.~\ref{fig: compare_model}.
Ultimately, these results confirm a reciprocal reinforcement: generative trajectories reconstructed as interpretive signals refine semantic boundaries, which in turn guide higher-fidelity synthesis, allowing {\modelname} to significantly improve generation while preserving its core multimodal intelligence.

\subsubsection{Model Architecture}
We first demonstrate the effectiveness of our method on BAGEL, where understanding and generation components are decoupled. To evaluate its generalization to tightly coupled architectures, we conduct a base model ablation on the purely autoregressive Janus-Pro-7b~\cite{januspro2025}.
Tab.~\ref{tab: model_ablation} shows that our method improves Janus by +3.2 on TIIF, +7.0 on WISE, and +4.7 on OneIG-EN, with the most pronounced gain on the knowledge-intensive WISE benchmark. This suggests that the proposed approach enhances knowledge expression by guiding generation through improved understanding, a mechanism that generalizes across different model architectures.
\subsubsection{Scaling Law for {\modelname}}

Scaling laws guide architectural design and optimization~\cite{kaplan2020scaling, chen2024agent}, but prior methods scale poorly due to reliance on external models or fixed prompts. In contrast, {\modelnamec} achieves scalable self-improvement through unbounded self-sampling and efficient Cognitive Pattern Reconstruction.
To explicitly assess scalability, we conduct scale-up experiments by varying the amount of self-generated data across \{1k, 5k, 8k, 10k, 20k\}.
\input{tables/model_abla}

\begin{figure}[t]
    \centering
    \includegraphics[width=1.0\columnwidth]{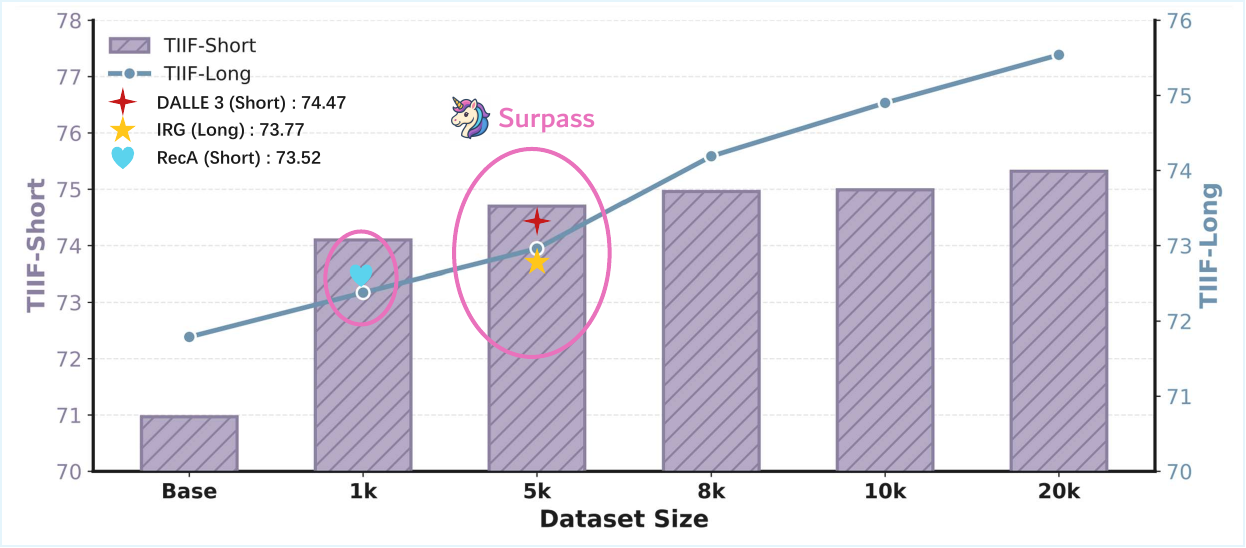}
        \vspace{-2em}
      \caption{\textbf{Data scaling result on TIIF}. The score consistently improves when the dataset size scales up. Notably, {\modelname} surpasses many powerful models only using {5k} training data.}
    \vspace{-2em}
    \label{fig:data_scale}
\end{figure}

As shown in Fig.~\ref{fig:data_scale}, with only 1K training samples, our method already surpasses RecA ~\citep{xie2025reconstruction} on TIIF. As the data scale increases, the model’s generative performance continues to improve, with more pronounced gains on long-prompt generation; notably, with just \textbf{5k} samples, it outperforms IRG~\cite{interleaving-reasoning} trained on \textbf{30k} GPT-4o distilled data as well as the strong closed-source model DALL·E 3~\cite{betker2023improving}.
These results reveal a favorable scaling regime in which self-generated data alone suffices to drive continual and efficient improvements in generative capability.

\subsection{Analysis}
We design a series of experiments for in-depth analysis to address the following two questions.

\noindent\textbf{Q1: Is self-play necessary?}

For self-play assessment, we use Qwen3-VL-235B-A22B-Instruct~\cite{yang2025qwen3} for data construction (UniCorn*). As shown in Tab.~\ref{tab: model_ablation}, employing stronger proposers/judges yields diminishing returns, where high costs and training time outweigh performance gains. This likely stems from the difficulty of fitting high-entropy teacher distributions, increasing latency without proportional information gain. We then compare {\modelname} with four unified models on \evalname{} (Judge: Qwen3-235B-A22B~\cite{yang2025qwen3}). \evalname{} requires both generation and understanding, reducing task bias and evaluating the model's self-reflection, thus signaling comprehensive multimodal intelligence.

As shown in Tab.~\ref{tab:hard-scores}, {\modelname} achieves the highest Hard score (46.5), outperforming its base BAGEL by nearly 10 points and others by over 3 points. UniCorn* lags by 6.5 points, suggesting strong external supervision yields disproportionate costs and insufficient unified coordination. This demonstrates that self-play enhances unified capabilities by distilling understanding into generation without degradation, achieving SOTA on \evalname. In contrast, Janus-Pro significantly underperforms comparable-scale models on \evalname{}, revealing a gap between its generation and self-understanding.

\noindent\textbf{Q2: Why {\modelname} works?}

{\modelname} addresses the asymmetry in Unified Multimodal Models where strong understanding capabilities remain inactive during generation. We identify three critical limitations: (1) the model lacks a holistic perception of the content it is about to generate, (2) it does not actively assess the quality of its own outputs during generation, and (3) it lacks the ability to reflect on and revise suboptimal generations. UniCorn resolves this by enabling understanding to supervise generation through a unified self-improvement loop involving captioning, evaluation, and reflection, restoring alignment for more faithful and controllable results.

Theoretically, we justify this approach using Mutual Information and Bayes' theorem, demonstrating that our task decomposition effectively minimizes Negative Log Likelihood (NLL). This guarantees that the auxiliary understanding tasks mathematically optimize the final unified objective. Detailed derivation is presented in Appendix \ref{appd:theoretical_analysis}.

%% file: tables/data_ablation.tex
\begin{table*}[t]
\centering

% Left: Table 2 (ablation)
\begin{minipage}[t]{0.74\textwidth}
\vspace{0pt}
\centering
{\renewcommand{\arraystretch}{0.9}%
\setlength{\tabcolsep}{1.5pt}%
\small
\begin{tabular}{lcc|cccccc}
\toprule
\textbf{Setting} & \textbf{TIIF-S} & \textbf{TIIF-R} & \textbf{MME-P} & \textbf{MME-C} & \textbf{MMB} & \textbf{MMMU} & \textbf{MMVP} & \textbf{MMStar} \\
\midrule
Base     & 71.0 & 70.7 & \textbf{1685.0} & \underline{696.0} & \textbf{84.6} & 52.8 & 69.3 & \textbf{65.0} \\
    Ours     & \textbf{74.7} & \textbf{78.4} & 1660.0 & 677.0 & 84.1 & \textbf{53.8} & \underline{70.0} & \textbf{65.0} \\
    w.o. CJR & 72.3 & 74.0 & 311.0 & 92.0 & 24.3 & 23.0 & 7.10 & 21.0 \\
    w.o. R   & 73.8 & 75.9 & 1632.0 & 655.0 & 84.2 & \underline{53.3} & \textbf{71.3} & \textbf{65.0} \\
    w.o. J   & 74.2 & 74.8 & 1542.0 & 478.0 & 82.6 & 51.9 & 65.3 & 61.0 \\
    w.o. C   & \underline{74.5} & \underline{76.4} & 1653.0 & \textbf{701.0} & \underline{84.3} & 50.9 & 68.0 & 64.0 \\
    w.o. G   & 73.4 & 72.3 & \underline{1669.0} & 685.0 & 84.2 & 53.0 & \underline{70.0} & 64.0 \\\bottomrule
\end{tabular}
}
\captionof{table}{\textbf{Ablation study on data composition}. Each variant is trained independently by removing exactly one data type from the full GCJR setting while keeping all other components fixed. S and R denote Short Score and Real Score, respectively.}
\label{tab:ablation_gcjr}
\end{minipage}
\hfill
% Right: Table 4 (hard score)
\begin{minipage}[t]{0.24\textwidth}
\vspace{0pt}
\centering
{\renewcommand{\arraystretch}{0.9}%
\footnotesize
\setlength{\tabcolsep}{5pt}
\begin{tabular}{@{}lc@{}}
\toprule
\textbf{Model} & \textbf{Hard score} \\
\midrule
Bagel & 36.6 \\
Show-o2 & 36.1 \\
Janus-Pro & 9.9 \\
UniCorn* & \underline{40.0} \\
{\modelname} & \textbf{46.5} \\
\bottomrule
\end{tabular}
}
\captionof{table}{\textbf{Hard score results on {\evalname}.} Soft score results are reported in Appendix~\ref{tab:soft-scores}.}
\label{tab:hard-scores}
\end{minipage}

\end{table*}

%% file: tables/model_abla.tex
\begin{table}[t]
\small
\renewcommand{\arraystretch}{1.4}
\setlength{\tabcolsep}{5pt}
\centering
\begin{tabular}{lccc}
\toprule
\textbf{Model} & {\textbf{TIIF}\phantom{\scriptsize+0.0}} & {\textbf{WISE}\phantom{\scriptsize(+0.0)}} & {\textbf{OneIG-EN}\phantom{\scriptsize+0.0}} \\
\hline
% 在第一行加入占位符，让它的宽度和第二行一致
Janus Pro & 63.2\phantom{\scriptsize+0.0} & 35.0\phantom{\scriptsize+0.0} & 26.7\phantom{\scriptsize+0.0} \\
+UniCorn  
& 65.9\textbf{\textcolor{ForestGreen}{\scriptsize+2.7} }
& 42.0\textbf{\textcolor{ForestGreen}{\scriptsize+7.0}} 
& 31.4\textbf{\textcolor{ForestGreen}{\scriptsize+4.7}} \\
\hline
UniCorn  & 73.8\phantom{\scriptsize+0.0} & 55.0\phantom{\scriptsize+0.0} & 42.6\phantom{\scriptsize+0.0} \\
UniCorn*
& 74.4\textbf{\textcolor{ForestGreen}{\scriptsize+0.6} }
& 54.0\textcolor{red}{\scriptsize-1.0} 
& 44.9\textbf{\textcolor{ForestGreen}{\scriptsize+2.3}} \\
\bottomrule
\end{tabular}
\caption{\textbf{Ablation studies on the base model (top) and the self-play framework (bottom)}. Unicorn$^{*}$ denotes the BAGEL model trained on data constructed using Qwen3-VL-235B-A22B-Instruct.}

\label{tab: model_ablation}
\end{table}

%% file: sections/5_conclusion.tex
\section{Conclusion}
In this paper, we propose {\modelnamec}, a self-supervised post-training framework that unifies multimodal comprehension and generation within a single model via multi-agent self-play and Cognitive Pattern Reconstruction, distilling internal latent knowledge into high-quality generative signals without external supervision. Extensive experiments, including the UniCycle cycle-consistency benchmark, demonstrate significant improvements in T2I generation while preserving multimodal intelligence, highlighting self-contained feedback loops as a scalable path for unified multimodal models.

%% file: sections/99_appendix.tex
\clearpage
\appendix
\section*{Content} 
Due to space limitations, we present additional details, along with quantitative and qualitative results of our {\modelnamec}, in the appendix. The outline is provided below.
\begin{itemize}
    \item \textbf{A. Additional Details (Appendix \ref{appd:training_detail})}
    \begin{itemize}
        \item Details of Training Data
        \item Training Setup
        \item Details of T2I Benchmarks
    \end{itemize}
\item \textbf{B. More Related Work (Appendix \ref{appd:relate work})}
    \begin{itemize}
        \item Unified Multimodal Models 
        \item Self-Improvement for LLMs 
        \item Multi-Agent Systems for LLMs
        \item LLM-as-a-Judge
    \end{itemize}
    \item \textbf{C. Theoretical Analysis (Appendix \ref{appd:theoretical_analysis})}
    \begin{itemize}
        \item Bi-Directional Mutual Information
        \item Internalized Preference Judgement
        \item Trajectory of Self-Reflection 
        \item Objective Decomposition for Unified Multimodal Learning

    \end{itemize}
    \item \textbf{D. Benchmark Details (Appendix \ref{sec:unicycle details})}
    \begin{itemize}
        \item Data Construction
        \item Evaluation Prompt
        \item More Results

    \end{itemize}
    \item \textbf{E. Additional Results (Appendix \ref{appd:additional results})}
    \begin{itemize}
        \item Quantitative Results
        \item Qualitative Results
        \item Failure Cases

\end{itemize}
\end{itemize}
% \ref{appd:training_detail}
% \ref{appd:relate work}
% \ref{appd:theoretical_analysis}
% \ref{sec:unicycle details}
% \ref{appd:additional results}
% \ref{sec: llm usage}

\section{Experiment Details} \label{appd:training_detail}
\paragraph{Data}
The specific prompt category, judgement rubrics for \S~\ref{sec:roles} is shown in Tab~\ref{tab:t2i_rules}. We set different random seeds and $\text{cfg\_text\_scale}$ in image sampling for a single prompt, to increase diversity for better images. Moreover, to ensure data quality, we filter the groups of samples when the highest score produced by the Judge is less than a fixed threshold (we choose 7). 

For \S~\ref{sec:re-data}, we use dozens of hand-written templates, without increasing any computational complexity. For Caption data, note that some image generation prompts contain phrases like "generate an image" or "create an image. For these types of data, we transfer the traditional caption task into "reconstruct image generation prompt, serving as a generalized caption task, which enhances data diversity and maintains data quality. 

The data mixture we use is 5k for Generation, 5k for Caption, 3k for Judgement, and 1k for Reflection. 
The detailed examples for training data are shown in Tab.~\ref{tab:train_data_example}.

\paragraph{Training} 
We conduct the post-training phase using the AdamW optimizer ($\beta_1 = 0.9, \beta_2 = 0.95, \epsilon = 10^{-15}$) with a constant learning rate of $1 \times 10^{-5}$. To ensure training stability, we implement a warm-up of 50 steps and apply gradient clipping at a threshold of 1.0. We set the total training duration to 600 steps, utilizing a gradient accumulation of 4 steps to manage the effective batch size. Furthermore, we apply an EMA ratio of 0.99 and balance the training objective with a CE to MSE loss weight ratio of $0.1:1$. For task-specific configurations, we utilize a maximum context window of 40k tokens, with generation and understanding resolutions set to $(512, 1024)$ and $(378, 980)$, respectively. Finally, a diffusion timestep shift of 4.0 is applied to calibrate the generative process.
We conduct all experiments on 8 NVIDIA H800 (80 GB) GPUs. Tab.~\ref{tab:training_recipe} shows the detailed hyperparameter configurations when post-training BAGEL.

\paragraph{Benchmarks}
To evaluate the generative performance of our model, we employ six representative text-to-image (T2I) benchmarks that assess various dimensions of synthesis quality and semantic alignment:

\begin{itemize} \item \textbf{TIIF}: This benchmark evaluates the model's ability to follow complex prompts across different lengths, specifically categorized into TIIF-S (short) and TIIF-L (long) to measure fine-grained text-to-image alignment. 
We use the official testmini subset and choose QwenVL2.5-72B~\cite{bai2025qwen2} as the evaluators.
\item \textbf{WISE}: This metric focuses on spatial consistency and visual fidelity, utilizing normalized scores to reflect the model's performance in complex scene layout generation. \item \textbf{OneIG}: A large-scale generative benchmark designed to test the robustness and diversity of the model across a wide array of semantic categories. \item \textbf{CompBench}: This benchmark targets compositional generation, specifically assessing how well the model handles attribute binding, object relations, and numerical constraints. \item \textbf{DPG} : DPG emphasizes the reconstruction of dense, multi-entity prompts, requiring the model to accurately synthesize multiple subjects and their respective fine-grained attributes. \item \textbf{GenEval}: A comprehensive evaluation framework that employs automated metrics to quantify core generative capabilities, including object recognition and attribute alignment. \end{itemize}

Together, these benchmarks provide a multi-dimensional assessment of our model's capacity to transform abstract conceptual prompts into high-fidelity visual outputs while maintaining strict adherence to textual constraints.

\begin{table*}[ht]
    \centering
    % \caption{\textbf{Training recipe of \modelname.}}
    % \label{tab:training_recipe}
    \renewcommand{\arraystretch}{1.1} % 稍微增加行高
    \begin{tabular}{l|c}
        \toprule
        \textbf{Hyperparameters} & \textbf{Post-training} \\
        \midrule
        Learning rate & $1\times10^{-5}$ \\
        LR scheduler & Constant \\
        Weight decay & 0.0 \\
        Gradient norm clip & 1.0 \\
        Gradient accumulation steps & 4 \\
        EMA ratio & 0.99 \\
        Loss weight (CE: MSE) & 0.1: 1 \\
        Optimizer & AdamW ($\beta_1=0.9, \beta_2=0.95, \epsilon=10^{-15}$) \\
        Warm-up steps & 50 \\
        Training steps & 600 \\
        Max context window & 40k \\
        Gen resolution \small{(min short side, max long side)} & (512, 1024) \\
        Und resolution \small{(min short side, max long side)} & (378, 980) \\
        Diffusion timestep shift & 4.0 \\
        \bottomrule
    \end{tabular}
        \caption{\label{tab:training_recipe}\textbf{Training recipe of \modelname.}}

\end{table*}

\section{More Related Work}
\label{appd:relate work}
%1.umm的经典架构 bagel，emu，showo等+self improving的方法@wzh
%2.self-play/self-improving在cv/llm/vlm领域的进展@myc

\paragraph{Unified Multimodal Models}
UMMs aim to unify cross-modal understanding and generation, yet a persistent challenge is that strong understanding does not reliably translate into equally strong native generation. Most UMMs follow two main architectural routes. \emph{Pure autoregressive} models jointly predict text and visual tokens with a unified next-token objective over interleaved sequences, as in Janus-Pro, Emu, and MetaMorph~\cite{chen2025janus,cui2025emu3,tong2025metamorph}. \emph{Hybrid} designs keep autoregressive modeling for language while relying on diffusion-style synthesis for continuous images, either via integrated modeling within a single backbone (e.g., Show-o and MonoFormer~\cite{xie2024show,zhao2024monoformer}) or through modular routing and sparse experts (e.g., LMFusion, Mixture-of-Transformers, and BAGEL~\cite{shi2024lmfusion,liang2024mixture,deng2025emerging}), with related paradigms such as Diffusion Forcing exploring diffusion-style guidance for interleaved generation~\cite{chen2024diffusion,huang2025interleaving}. Recent work also explores image generation foundation models that systematize text-conditioning and training recipes on diffusion-style backbones. Qwen-Image~\cite{wu2025qwen} adopts a double-stream MMDiT, conditioned on a frozen Qwen2.5-VL encoder and a VAE tokenizer, and uses progressive/curriculum training with multi-task objectives to cover settings such as multilingual text rendering and editing. Z-Image~\cite{cai2025z} proposes a 6B single-stream diffusion Transformer (S3-DiT) and derives a Turbo variant via few-step distillation and reward post-training, focusing on inference efficiency under low sampling steps and text-rendering-related scenarios. Beyond architecture, recent work investigates self-improvement by turning self-produced signals into training objectives~\cite{yu2025guided,zhou2024calibrated,wang2025unified}. For UMMs, SRUM leverages internal understanding as an evaluator to derive fine-grained rewards~\cite{srum}, and UniRL couples generation and understanding via supervised and reinforcement learning~\cite{unirl}. Complementary directions also study data-centric enhancement for vision-language alignment, e.g., ultra-detailed caption generation to enrich training signals for VLMs~\cite{zeng2025enhancing}, and interaction-based learning setups that emphasize perception and reasoning behaviors~\cite{zeng2025agentic}. However, existing self-improvement pipelines often depend on auxiliary components or externally produced dense feedback, as well as task-specific reward shaping, fixed prompt pools, or pre-selected concepts, which can limit scalability and generalization when extending self-improvement to broader UMM settings.

\paragraph{Self-Improvement for LLMs}
Self-play drives autonomous improvement by pairing self-generated challenges with outcome-driven learning~\cite{silver2017mastering}. In LLMs, this enables zero-data self-evolution: models generate training signals without curated datasets, as exemplified by the uncertainty and self-consistency curricula of R-Zero~\cite{huang2025r} as well as the executor-verified rewards of Absolute Zero~\cite{zhao2025absolute}. Beyond task generation, self-produced evaluation guides preference learning and reasoning, through methods such as self-rewarding feedback~\cite{yuan2024self}, constraint-based optimization~\cite{zhou2024calibrated}, reflective reward learning~\cite{choi2024self}, and process-consistency rewards for long-horizon tasks~\cite{guo2025r1}. In multimodal settings, related efforts incorporate retrieval-augmented reasoning with reinforcement learning to improve understanding over visually rich information sources~\cite{wang2025vrag}. Extending to Unified Multimodal Models (UMMs), their integrated modules naturally enable self-improvement, with the understanding module providing internal multi-scale feedback to guide generation, establishing a promising paradigm for fully model-driven enhancement.

\paragraph{Multi-Agent Systems for LLMs}
LLM-based multi-agent systems instantiate role-specialized agents to decompose tasks, explore diverse solutions, and cross-check results, supported by orchestration frameworks such as AGENTVERSE~\cite{chen2024agentverse} and debate-style interactions that encourage diverse hypotheses and mutual critique~\cite{liang2024encouraging}. While many systems are primarily deployed at inference time, recent work explores closed-loop training with self-generated signals, including multi-agent training and self-play pipelines~\cite{motwani2024malt,zhao2025absolute,chen2025multi}. Beyond pure language agents, embodied and action-centric extensions seek to improve generalization by structuring the coupling between reasoning and action, as in DualVLA~\cite{fang2025dualvla}. However, empirical analyses show these systems can be brittle and costly, with recurring coordination and verification failures that limit scalability and generalization~\cite{cemri2025multi}. Motivated by these limitations, our work uses lightweight role instantiation within a single unified multimodal model, turns role interactions into self-training signals for improving native multimodal generation, and introduces a cycle-consistency benchmark to test whether gains reflect genuine multimodal understanding rather than task-specific tuning.

\paragraph{LLM-as-a-Judge}
Recent work increasingly adopts \emph{LLM-as-a-Judge} as a scalable alternative to human evaluation for open-ended generation and benchmark construction, where strong LLMs provide pointwise scores or comparative rankings with broader coverage than heuristic metrics at lower annotation cost~\cite{li2025generation,li2024llms}. However, LLM-based judges are not uniformly reliable. Their judgements can be sensitive to prompt phrasing and candidate presentation, and they may exhibit systematic biases and vulnerabilities, including adversarial manipulation~\cite{raina2024llm,thakur2025judging,li2024llms}. These concerns motivate meta-evaluation of judges and evaluation protocols that reduce reliance on fragile or implicit judgement signals~\cite{feng2025we}, as well as targeted benchmarks that stress-test self-critique in tool-calling error scenarios~\cite{huang2025critictool}. In multimodal evaluation, video reasoning benchmarks further broaden coverage beyond static images, including chain-of-thought video reasoning and visual-prompt-based interaction protocols~\cite{qi2025vcr,zhao2025v2p}. In our setting, judge models serve two roles. For T2I generation, we use a VLM-based judge with task-specific rubrics to assess prompt-image alignment and visual fidelity. For T2I2T evaluation, we use the same UMM as an LLM-based judge to verify whether predicted answers match the original instruction and reference answers, enabling structured scoring at scale.

\section{Theoretical Analysis}\label{appd:theoretical_analysis}

%理论分析
In this section, we present a thorough theoretical analysis to explain why {\modelname} works.
As discussed in \S~\ref{sec:re-data} and Appendix~\ref{appd:training_detail}, we construct the following four types of data:
\begin{itemize}
    \item \textbf{Generation Data ($G$)}: High-quality images sampled by the model and selected via a Best-of-N strategy.
    \item \textbf{Captioning Data ($C$)}: Constructed via a reverse process, where the best images and caption prompts serve as input to predict the original generation prompts.
    \item \textbf{Judgement Data ($J$)}: Self-evaluation outputs from the model, including Chain-of-Thought (CoT) reasoning and final scoring.
    \item \textbf{Reflection Data ($R$)}: Self-correction data taking suboptimal images and editing instructions as input to output the optimal images.
\end{itemize}
Ablation studies demonstrate that each data type contributes to both generation and understanding capabilities, fostering a truly unified model. Below, we provide a theoretical analysis of why these four synthetic data types synergistically enhance image generation.

For most unified multimodal model like BAGEL, parameters for generation and understanding are shared partially. The objective is to learn the joint distribution of Text $(T)$ and Image $(I)$, denoted as $\pi_{\star} (T, I)$. We train the model $\pi_{\theta} (T, I)$ to approximate the constructed data distribution $p(T,I), (T, I) \sim \mathcal D$, where $\mathcal{D}$ is the predefined dataset, by minimizing the unified loss function $\mathcal{L}_{Unified}$.

\subsection{Bi-Directional Mutual Information}
Most existing works focus solely on constructing $p(I \mid T)$ to enhance generation. However, our experiments show that this single-directional training leads to a collapse in understanding capabilities. We analyze this from the perspective of \textbf{Mutual Information}.

Consider the mutual information between image and text, $MI(I; T)$
\begin{equation}
    \begin{split}
        MI(I; T) &= H(I) - H(I \mid T) \\
                 &= H(T) - H(T \mid I)
    \end{split}
\end{equation}
Constructing only generation data $p(I \mid T)$ minimizes an upper bound on $H(I \mid T)$(the conditional cross-entropy/NLL). However, according to the equation above, one-way likelihood training provides no direct training signal for the other. The model fails to capture the dependency of $T$ given $I$, leading to the collapse of understanding capabilities. Due to parameter sharing, this representational deficiency results in sub-optimal generation performance.

By constructing \textbf{Captioning Data} $p(T \mid I)$ via a self-dual approach, we encourage bidirectional consistency between the two conditionals:
\begin{equation}
    \begin{split}
        p(I, T) &= p(I \mid T)\, p(T) \\
                &= p(T \mid I)\, p(I),\quad (T, I) \sim \mathcal{D}_C
    \end{split} 
\end{equation}
where $\mathcal{D}_C$ is the caption dataset we constructed, and $p(I)$, $p(T)$  are priors determined by both the dataset and model.

This explains why Caption data not only preserves understanding capabilities but also enhances generation by enforcing a more robust, bi-directionally consistent multimodal representation.

\subsection{Internalized Preference Judgement}

A truly unified multimodal model requires not only the ability to generate and understand but also the capacity to align with human preferences—specifically, the ability to \textbf{Judge}. We refine the target distribution to include judgement $J$
J, denoted as $\pi_{\star} (T,I,J)$. Using the chain rule of probability, the model's joint distribution can be decomposed as
\begin{equation}
    \pi_{\theta} (I, T, J) = \pi_{\theta}(J \mid I, T) \cdot \pi_{\theta}(I \mid T) \cdot \pi_{\theta}(T)
\end{equation}

This decomposition implies that the system is composed of text priors, text-to-image generation, and the ability to judge the quality of the $(I,T)$ pair. We construct judgement Data $p(J \mid I,T)$ to train the term $\pi_{\theta} (J \mid I,T)$. This allows the model to "internalize" evaluation metrics, effectively learning a discriminator that guides the generator toward higher-quality outputs.

\subsection{Trajectory of Self-Reflection }
With the introduction of judgement, the model can learn to improve from "bad" to "good" states. We aim for the model to generate the optimal image $I_w$ potentially via an iterative process.

Let $I$ denote a suboptimal image sampled during exploration. We can model the generation of the best image $I^*$ by introducing $I$ as an intermediate latent variable in the probability decomposition:

\begin{equation}
    \pi_{\theta} (I^* \mid T, J) =
    \pi_{\theta} (I^* \mid I, T, J) \cdot \pi_{\theta} (I \mid T, J)
\end{equation}

Here, $\pi_{\theta}  (I \mid T, J) $
represents the initial generation, and $\pi_{\theta}(I^* \mid I,T,J)$ represents the refinement step.By constructing \textbf{Reflection Data} $p(I^* \mid I,T,J)$, we explicitly train the model to act as a "correction operator". This enables the model to learn the trajectory of improvement, significantly boosting its ability to handle complex instructions and self-correction.

\subsection{Objective Decomposition for Unified Multimodal Learning}
From the perspective of  Negative Log-Likelihood (NLL), the overall loss function $\mathcal{L}_{Unified}$ for BAGEL is decomposed as follows:
\begin{equation}
    \mathcal{L}_{Unified} = \mathcal{L}_{G} + \mathcal{L}_{C} + \mathcal{L}_{J} + \mathcal{L}_{R}, 
\end{equation}
where $\lambda_{i}$ represents the relative data proportions across each dataset. The individual loss components are defined as:
\begin{equation*}
    \begin{aligned}
    \mathcal{L}_{G} &= - \mathbb{E}_{(I^*, T) \sim \mathcal{D}_{bon}} \left[ \log \pi_{\theta}(I^* \mid T) \right] \\[2ex] % 稍微拉大一点间距
    \mathcal{L}_{C} &= - \mathbb{E}_{(T, I^*) \sim \mathcal{D}_{C}} \left[ \log \pi_{\theta}(T \mid I^*) \right] \\[2ex]
    \mathcal{L}_{J} &= - \mathbb{E}_{(I, T, J) \sim \mathcal{D}_{J}} \left[ \log \pi_{\theta}(J\mid I, T) \right] \\[2ex]
    \mathcal{L}_{R} &= - \mathbb{E}_{(I^*, I, T, J) \sim \mathcal{D}_{R}} \left[ \log \pi_{\theta}(I^* \mid I, T, J) \right]
    \end{aligned}
\end{equation*}

where $\mathcal{D}_i$ represents different synthetic datasets.

\section{Benchmark Details}
\label{sec:unicycle details}
\subsection{Data Construction}
\begin{figure*}[h]
    \centering
    \includegraphics[width=\textwidth]{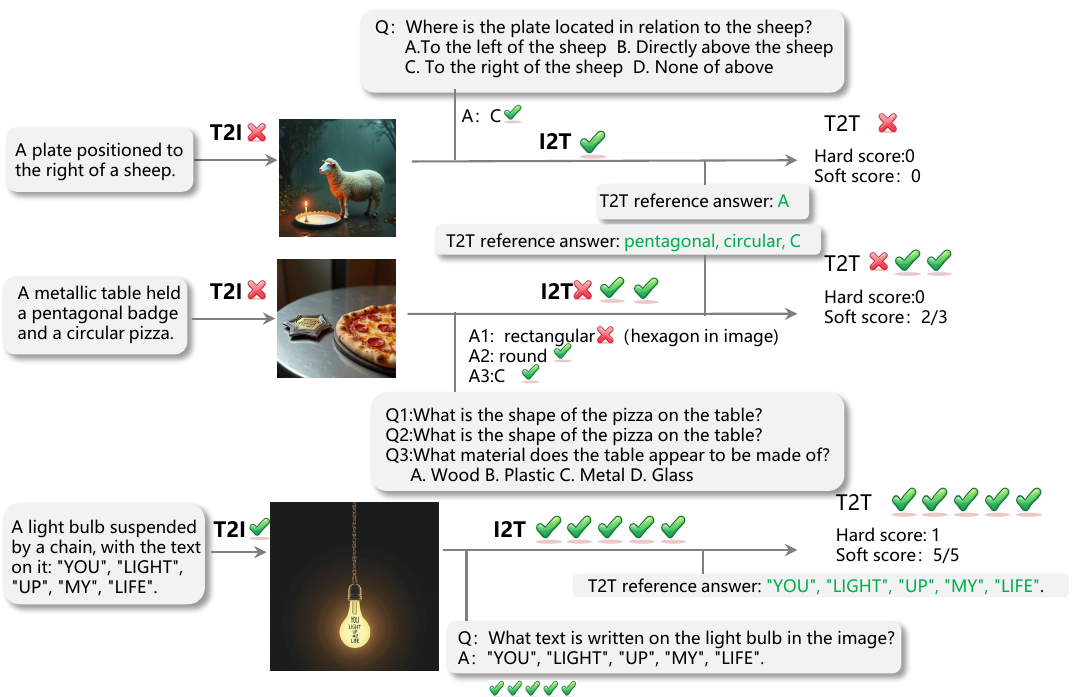}
    \caption{Cases of \evalname{}.}
    \label{fig:showcase}
\end{figure*}

Based on the TIIF benchmark, we generate question–answer pairs for instruction reconstruction, extending the original Text-to-Image (T2I) evaluation to a Text-to-Image-to-Text (T2I2T) setting. To balance task difficulty with answer evaluability, we design task- and question-type–specific prompt templates. For negation-related tasks, we construct prompts that elicit binary (yes/no) questions, ensuring unambiguous evaluation. For tasks like spatial relation, open-ended questions often lead to ambiguous judgments— for instance, an instruction specifies "left" but the generated image places an object in the "front-left" position, an answer such as "in front" may be plausible yet difficult to assess consistently. To improve evaluation stability, we therefore formulate these tasks as multiple-choice questions. For the remaining task types, such as color recognition and counting, we adopt open-ended question–answer formats to maintain sufficient difficulty and discriminative power. Moreover, we explicitly enforce task-type–based question completeness: since all instruction-implied information relevant to the task type is treated as a reconstruction target, a QA set is considered valid only if it fully covers the reconstruction targets without redundancy. QA pairs are generated using Qwen3-235B-A22B~\cite{yang2025qwen3} and subsequently annotated under a unified labeling protocol by experienced human annotators. After filtering, we retain 1,401 high-quality instances and totally 2968 questions ( The distribution of question types is shown in Tab.\ref{tab:question-distribution}) for evaluation, covering almost all task types present in the original TIIF benchmark. We present several cases in Fig.~\ref{fig:showcase}.

\subsection{{\evalname} Evaluation Prompt}
The prompt templates for T2I2T evaluation of \evalname{} are presented in Fig.~\ref{fig:evaluation}, \ref{fig:evaluation_text}.

\subsection{Soft scores results on {\evalname}}
Soft scores results of {\modelname} and the other four models on {\evalname} are shown in Tab.~\ref{tab:soft-scores}.

\section{Additional Results}
\label{appd:additional results}
\subsection{Quantitative Experiments} 
\label{sec: detail_results}
Detailed scores across the six T2I benchmarks are reported in Tab.~\ref{tab:oneig_en}, ~\ref{tab:tiif_qwen}, ~\ref{tab:wisescore},~
\ref{tab:compbench_final},~\ref{tab:geneval}and ~\ref{tab:dpg}.
We also evaluate our model on the image edit task \cite{rise} in Tab.~\ref{tab: edit}.
\input{tables/edit_result}
\subsection{Qualitative Results}
\label{sec: qualitative results}
The qualitative comparison of the reliance on external data and models between our approach and other methods is presented in Tab.~\ref{tab: model_compare}.
Without relying on external task-specific models or annotated data, UniCorn achieves state-of-the-art performance on OneIG-EN using only 5K training samples.
% Qualitative results comparing our method with RecA, SRUM, and variants using different data recipes are shown in Fig.~\ref{fig: compare_model}.
% More visualized cases of {\modelname} are provided in Fig~\ref{fig: vis}.

\subsection{Failure Cases}
\begin{figure*}[h]
    \centering
    \includegraphics[width=\textwidth]{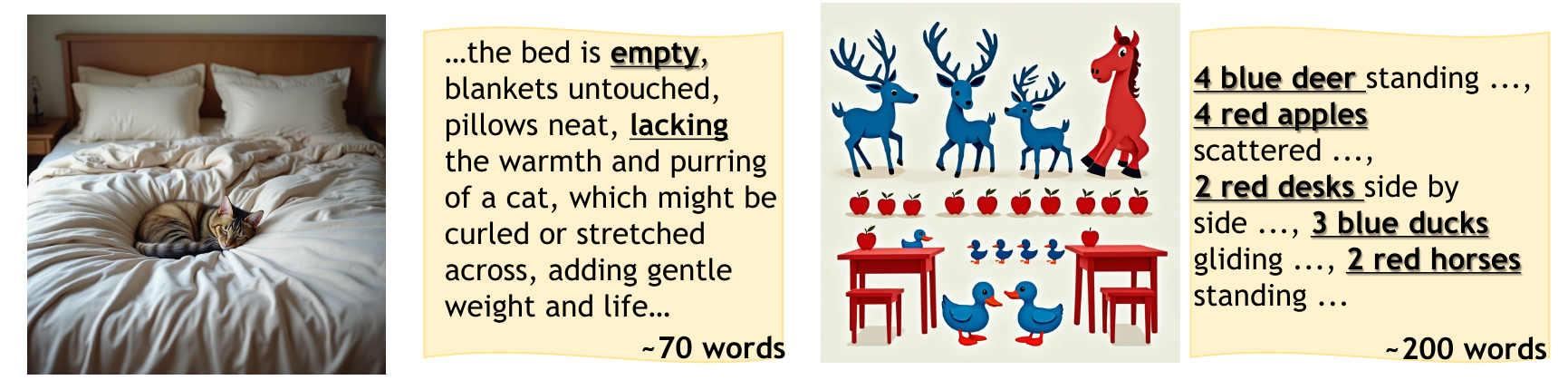}
    \caption{Failure cases of {\modelname} in chanllenging tasks of Negation and Counting.}
    \label{fig: fail cases}
\end{figure*}

In Fig.~\ref{fig: fail cases}, we show two failure cases of {\modelname} in challenging tasks such as Negation and Counting. We attribute the model's limitations on these tasks to their inherent difficulty for multimodal models. Within our self-play training paradigm, it is challenging to provide effective supervision for such tasks; consequently, the lack of significant improvement is consistent with our expectations.

\clearpage

\input{tables/T2I_rule}

\input{tables/train_data_examples}
\input{tables/detail_tables}
\input{tables/generation_prompt}
\input{tables/judge_prompt}

\input{tables/compare_self}

\input{tables/eval_bench_prompt}
\input{tables/eval_bench_prompt_text}

\label{sec:appendix}

%% file: tables/edit_result.tex
\begin{table}[t]
  \centering
  \begin{tabular}{cc} % Three centered columns
    \toprule
    \textbf{Model} & \textbf{RISE Score} \\ % First row bold
    \midrule
    BAGEL & 33.33  \\
    \textbf{UniCorn} & 38.87(+5.54)  \\ % Last row bold
    \bottomrule

  \end{tabular}    
  \caption{The evaluation results of RISE.}
    \label{tab: edit}
\end{table}

%% file: tables/T2I_rule.tex
\begin{table*}[t]
\centering
\small
\renewcommand{\arraystretch}{1.3}
\begin{tabular}{p{2.6cm} p{4.2cm} p{4.2cm} m{3.2cm}}
\toprule
\textbf{Major Category} &
\textbf{Generation Requirement} &
\textbf{Judgement Rubrics} &
\textbf{Example} \\
\midrule
General Object &
Depict specific real-world objects or scenes, focusing on attributes including shape, color, texture, and single/multi-object composition. & 
Object existence, attribute accuracy (color/shape/texture), and compositional correctness. &
\includegraphics[height=1.6cm]{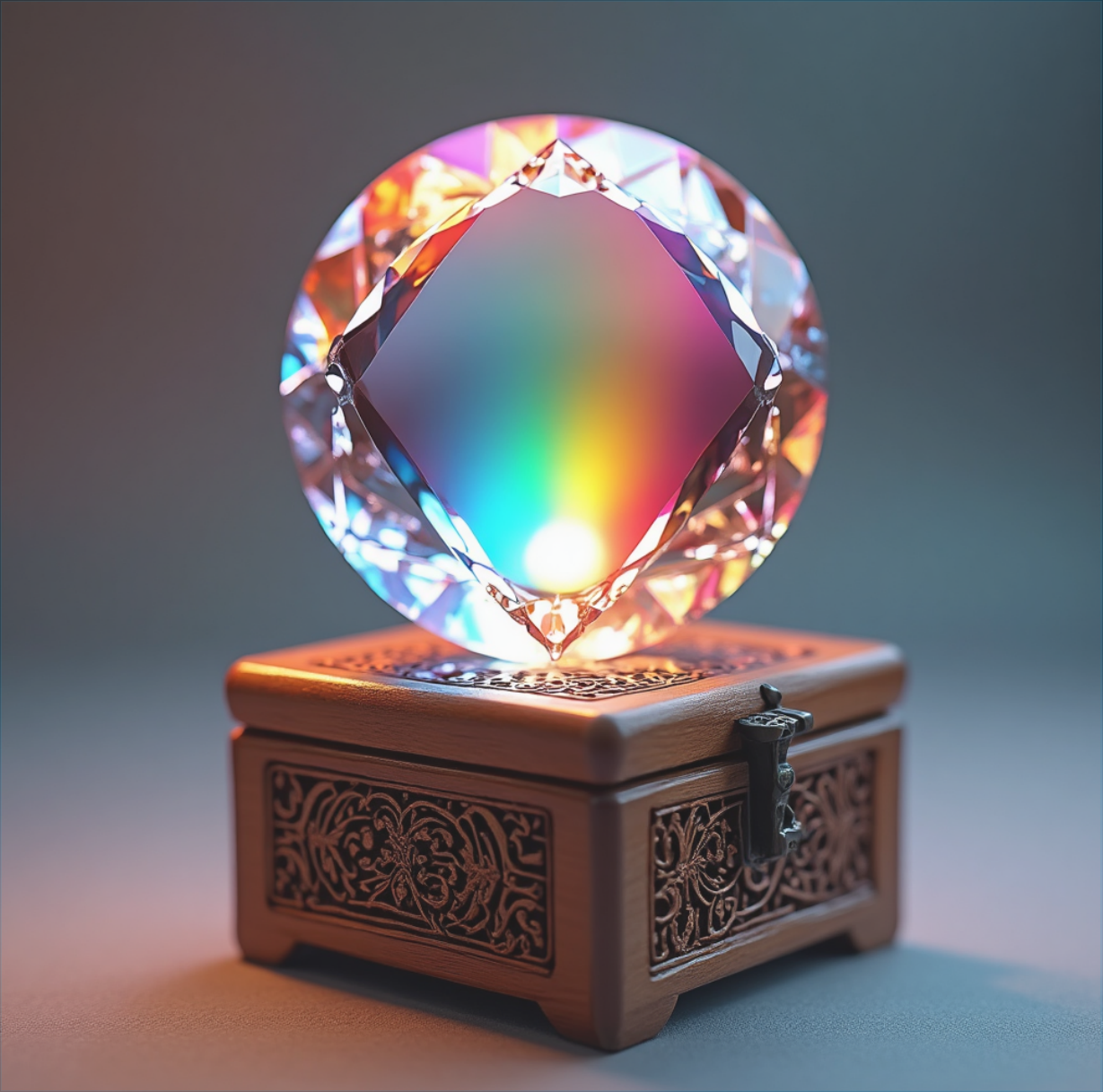} \\

Object Relations &
Reflect logical connections between objects, involving action \& interaction, comparison, differentiation, or negation. &
Logical correctness of relations (e.g., A is interacting with B), in addition to basic object correctness. &
\includegraphics[height=1.6cm]{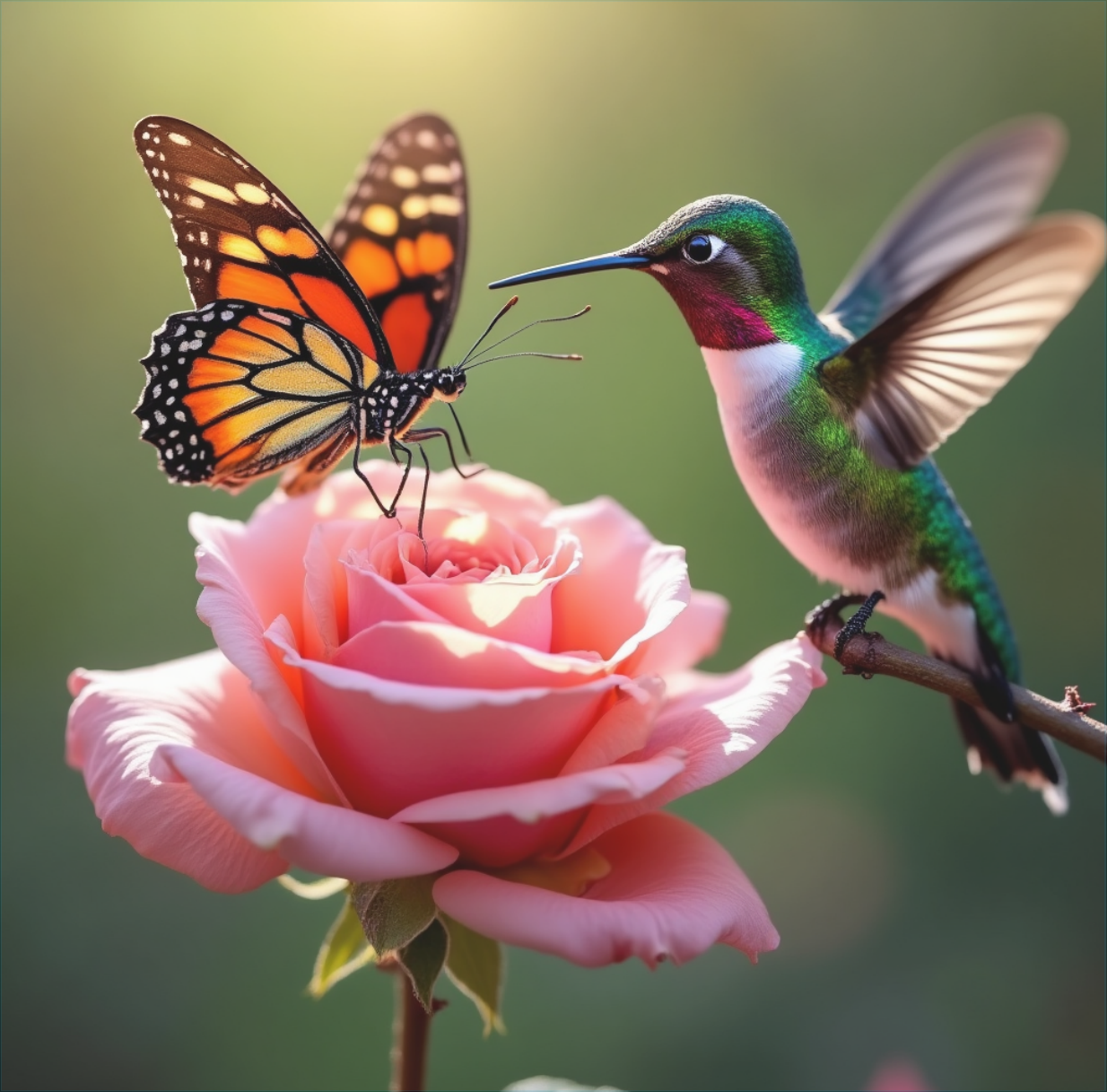} \\

General Knowledge &
Depict specific general elements requiring external knowledge in real life, such as festivals, sports, celebrities, religions, or crafts. &
Factual accuracy, cultural recognition, and attribute alignment with real-world entities. &
\includegraphics[height=1.6cm]{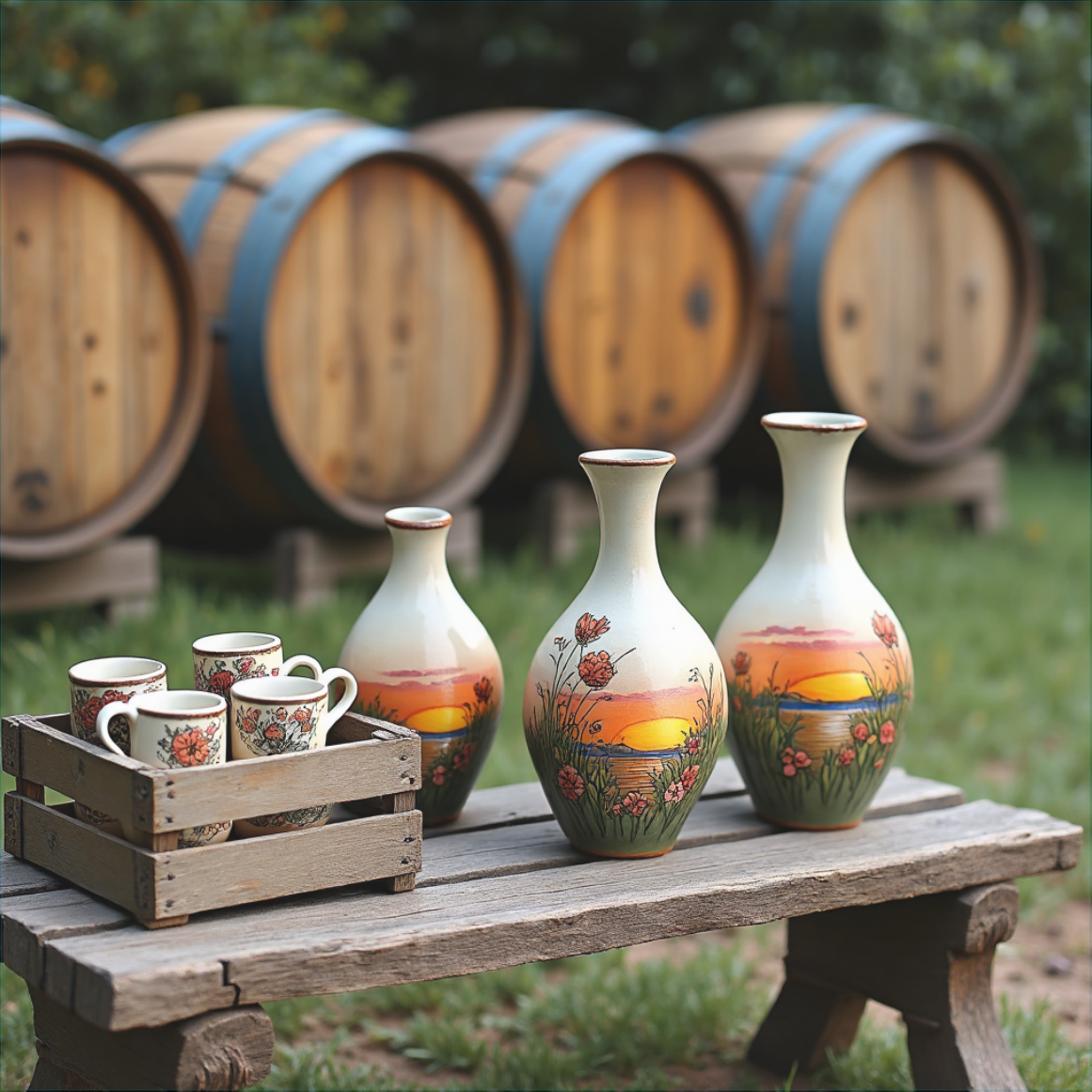} \\

% 4. Spatio Reasoning (对应原始类别 4)
Spatio Reasoning &
Handle complex spatial layouts, including 2D/3D structures, occlusion reasoning, and specific viewing perspectives (e.g., bird's-eye view). &
Spatial consistency, perspective correctness, and accurate handling of occlusions/depth. & 
\includegraphics[height=1.6cm]{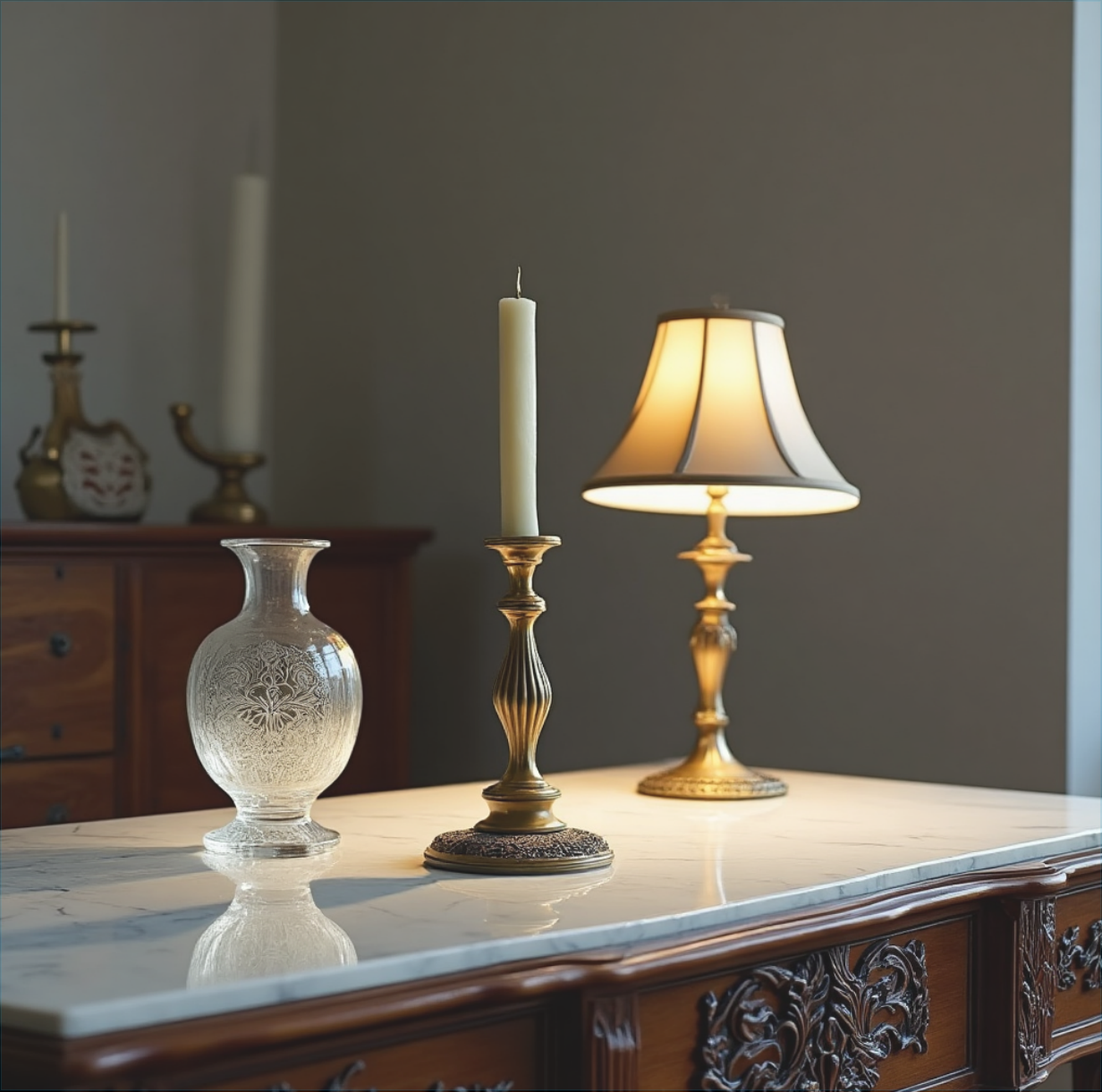} \\

% 5. Temporal Reasoning (对应原始类别 5)
Temporal Reasoning &
Reflect time-sensitive states, such as horizontal time (synchronous elements) or longitudinal time (chronological changes/stages). &
Temporal consistency, logical progression of states, and accuracy of time-specific features. & 
\includegraphics[height=1.6cm]{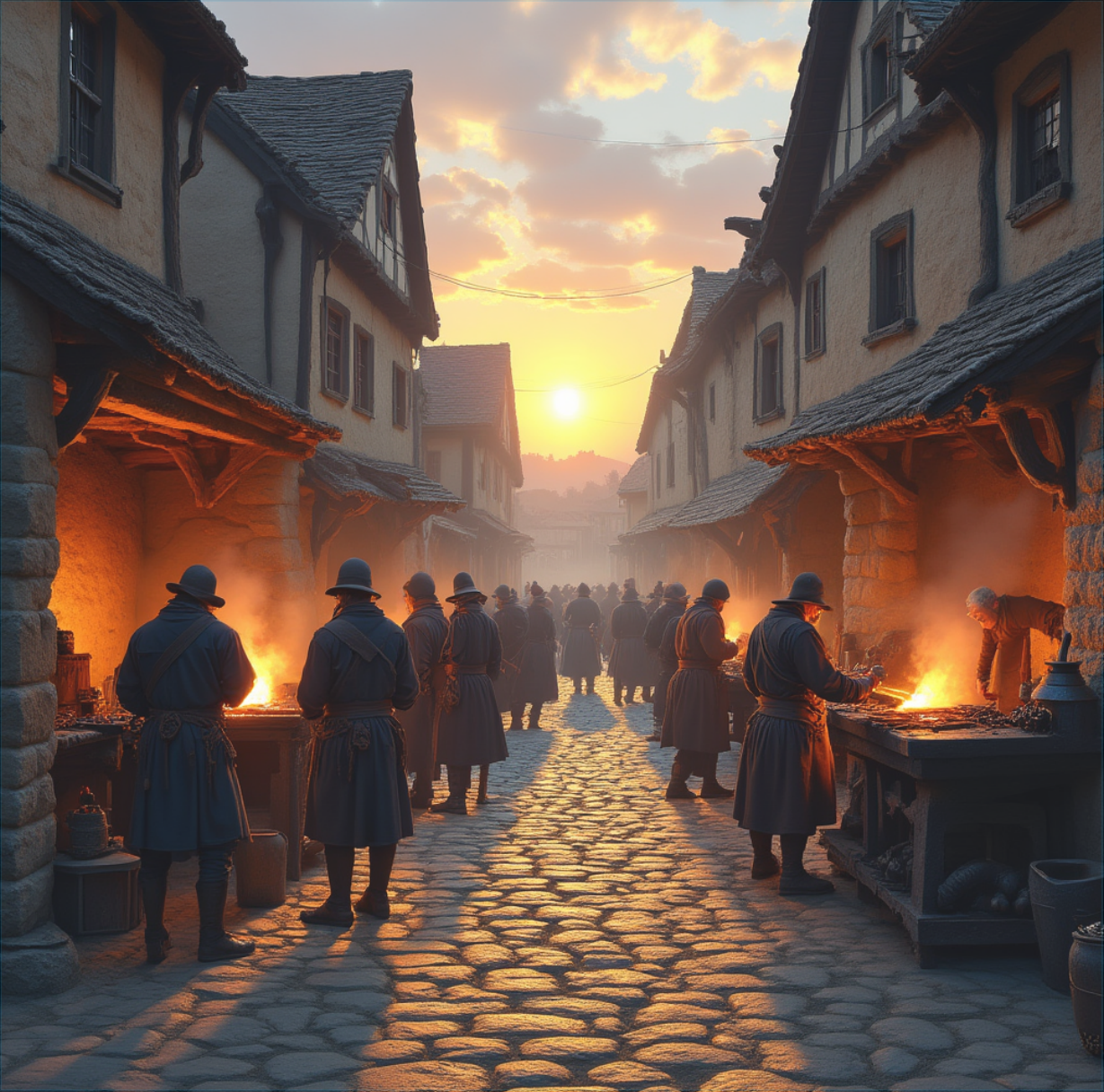} \\

% 6. Text Rendering (对应原始类别 8)
Text Rendering &
Render legible text within images across various formats: natural-scene text, designed posters/menus, or handwriting/graffiti. &
OCR accuracy (spelling), font style appropriateness, and text-background integration. & 
\includegraphics[height=1.6cm]{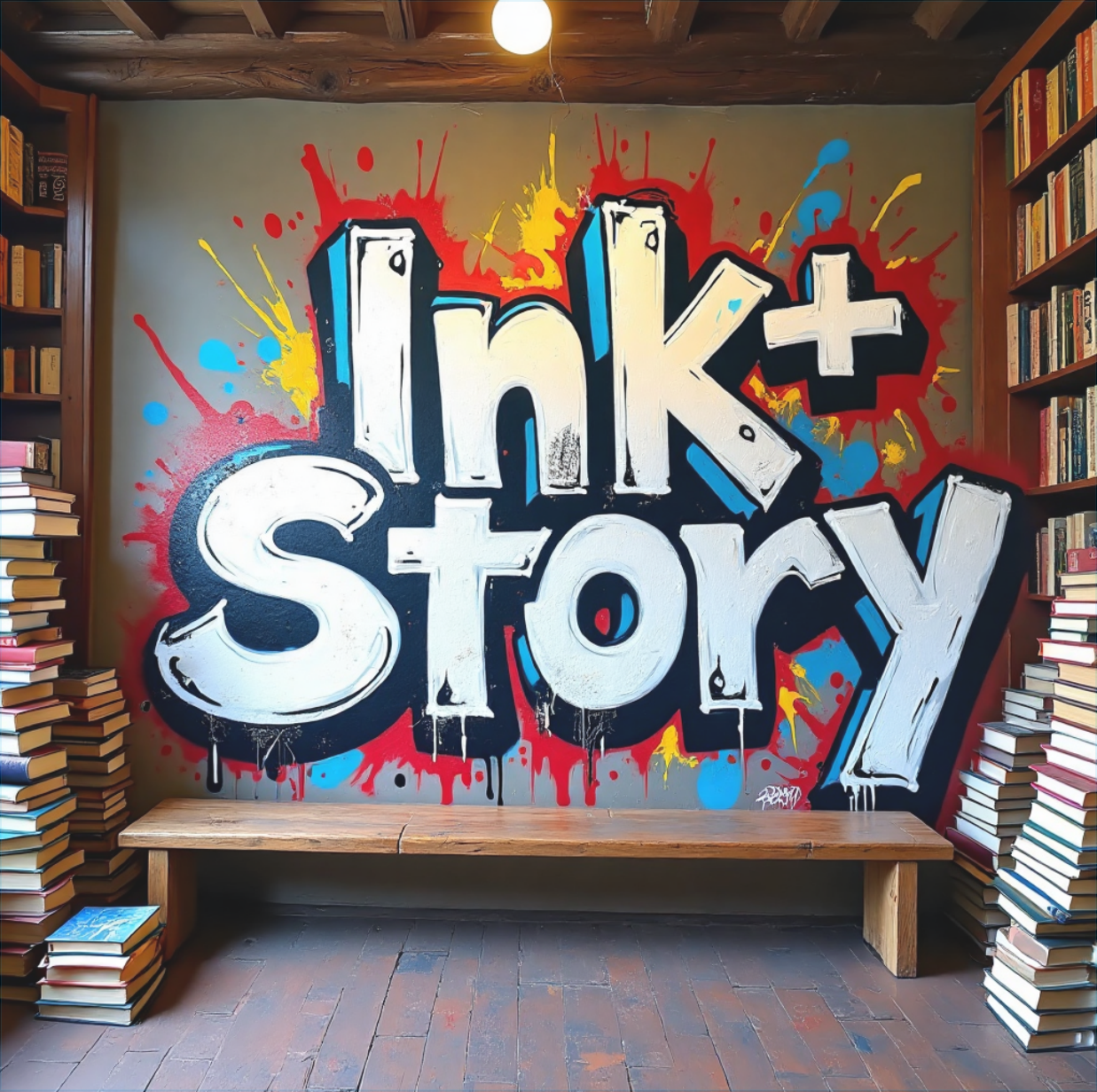} \\

% 7. Natural Science (对应原始类别 2)
Natural Science &
Accurately represent scientific subjects, including precise anatomy of animals/plants and physics/chemistry phenomena. & 
Scientific realism, biological anatomical correctness, and physical plausibility. &
\includegraphics[height=1.6cm]{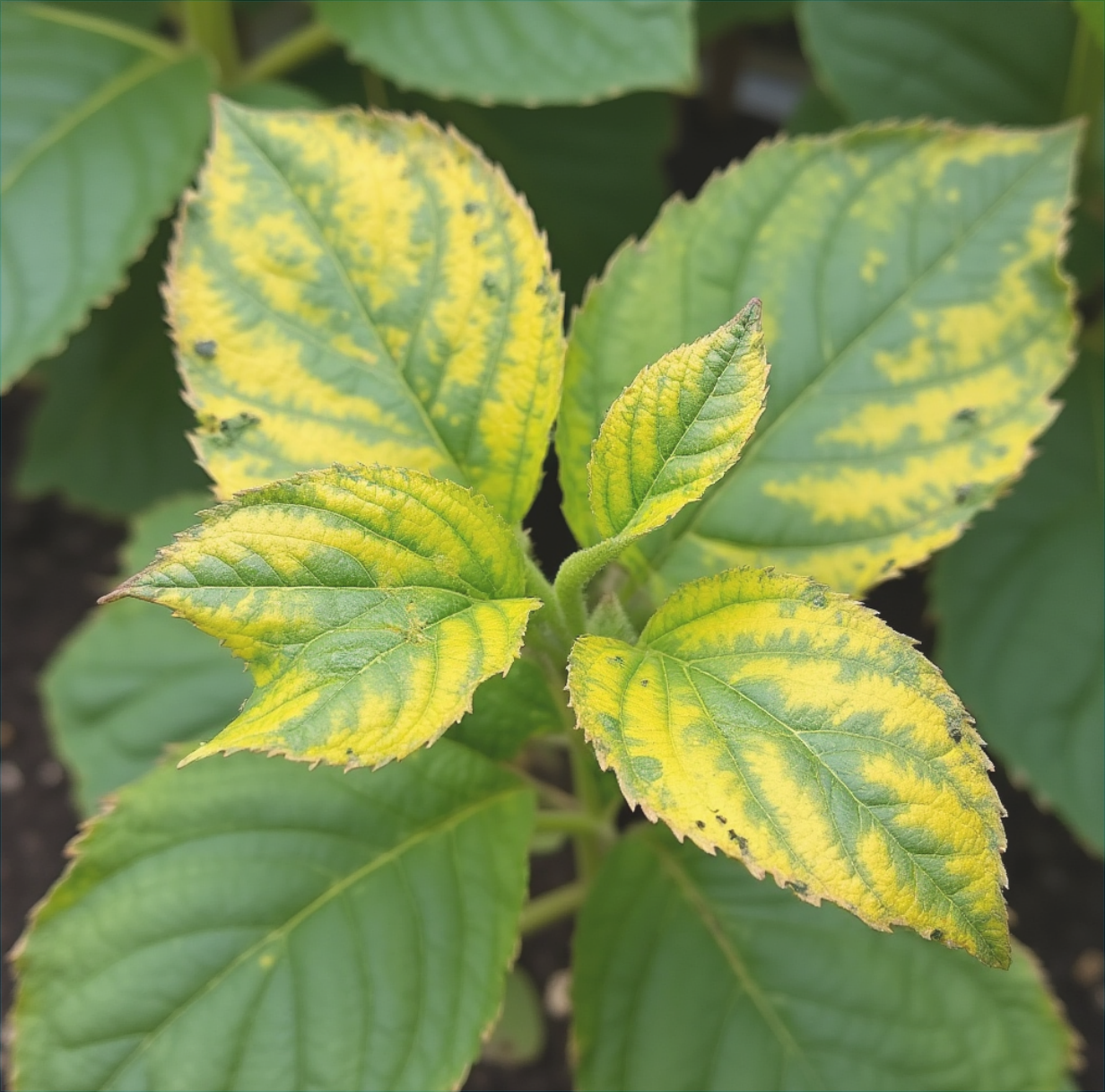} \\

% 8. Portrait (对应原始类别 7)
Portrait &
Generate human-centered portraits with specific framing requirements: close-up, half-body, or full-body shots. &
Framing accuracy (shot scale), facial/anatomical correctness, and identity consistency. &
\includegraphics[height=1.6cm]{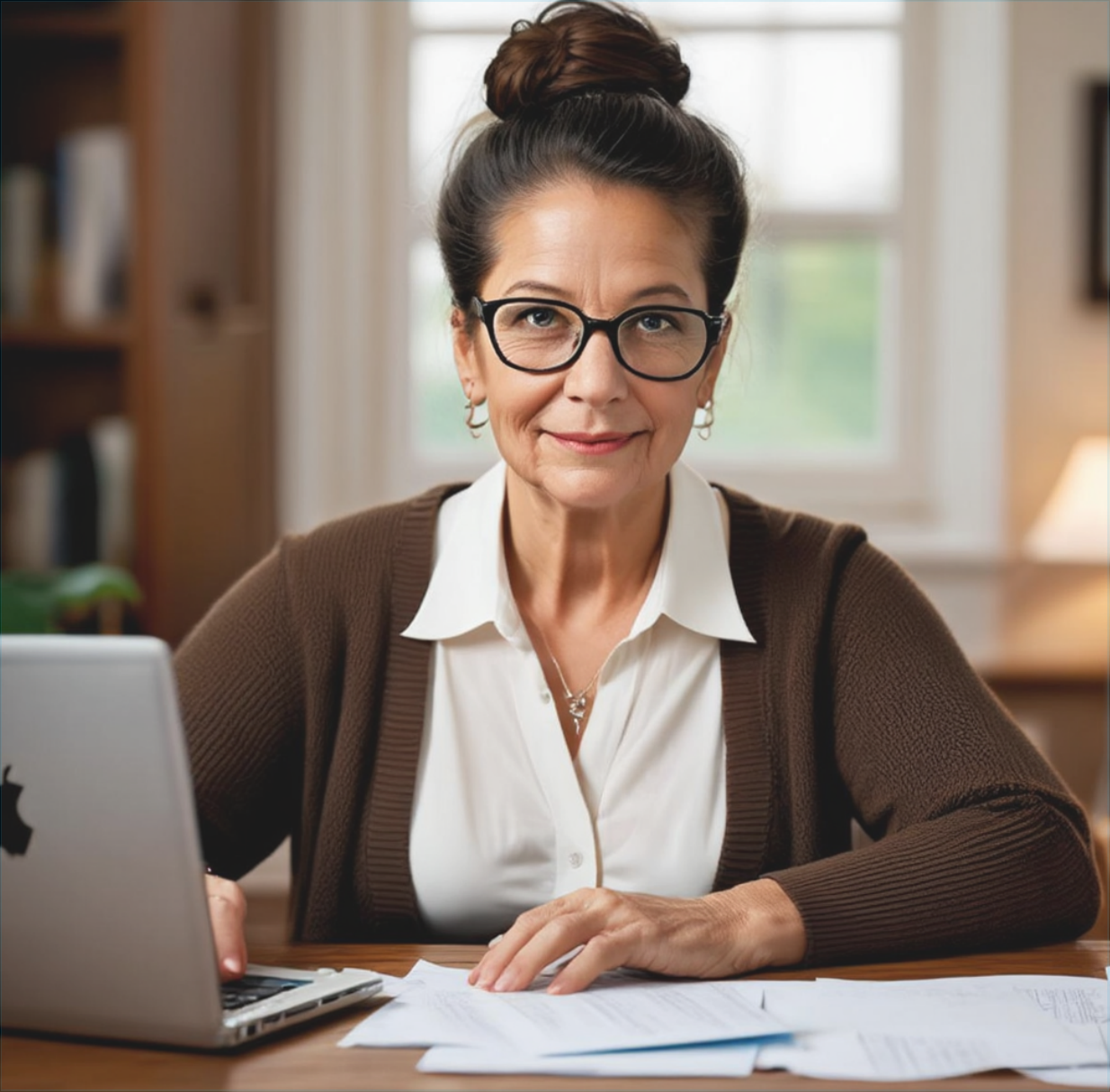} \\

% 9. Stylization (对应原始类别 9)
Stylization &
Adhere to specific artistic styles, primarily focusing on Anime style or various artistic stylizations (e.g., oil painting, sketch). &
Style fidelity, aesthetic quality, and texture application consistent with the requested art medium. &
\includegraphics[height=1.6cm]{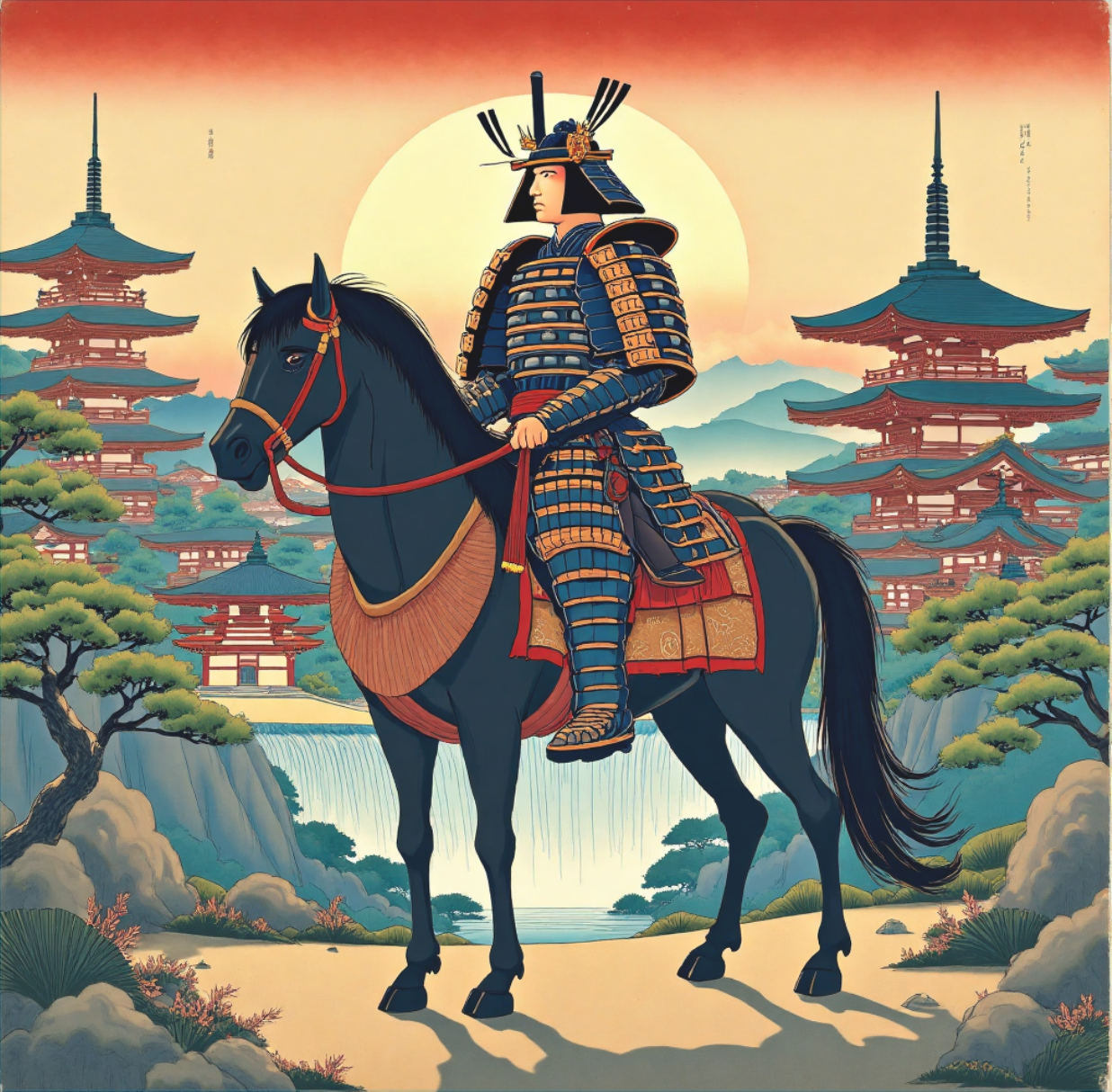} \\

% 10. Counting (对应原始类别 10)
Counting &
Generate a precise number of specific objects as described in the prompt. &
Count accuracy (numerical precision) and object distinctness. &
\includegraphics[height=1.6cm]{figures/figures_appendix/counting_cropped.pdf} \\

\bottomrule
\end{tabular}
\caption{Detailed data type range, description and judgement rubrics.}
\label{tab:t2i_rules}
\end{table*}

%% file: tables/train_data_examples.tex
% \begin{document}
\begin{table*}[htbp]
    \centering
    \renewcommand{\tabularxcolumn}[1]{m{#1}} 
    % 定义两列：l 代表左对齐。如果需要内容居中可以用 c
    \begin{tabularx}{\linewidth}{l X X}
        \toprule
        \textbf{Category} &\textbf{Prompt Example} & \textbf{Response} \\
        \midrule
        % 第一行: Generation
        % valign=c 让图片与文字垂直居中
        \textbf{Generation} & A glass sculpture in the shape of a turtle with intricate patterns of red lines on its shell, resting on a black marble pedestal, with soft light coming from above, highlighting the contours of the turtle and casting delicate shadows on the floor. & \includegraphics[width=0.66\linewidth]{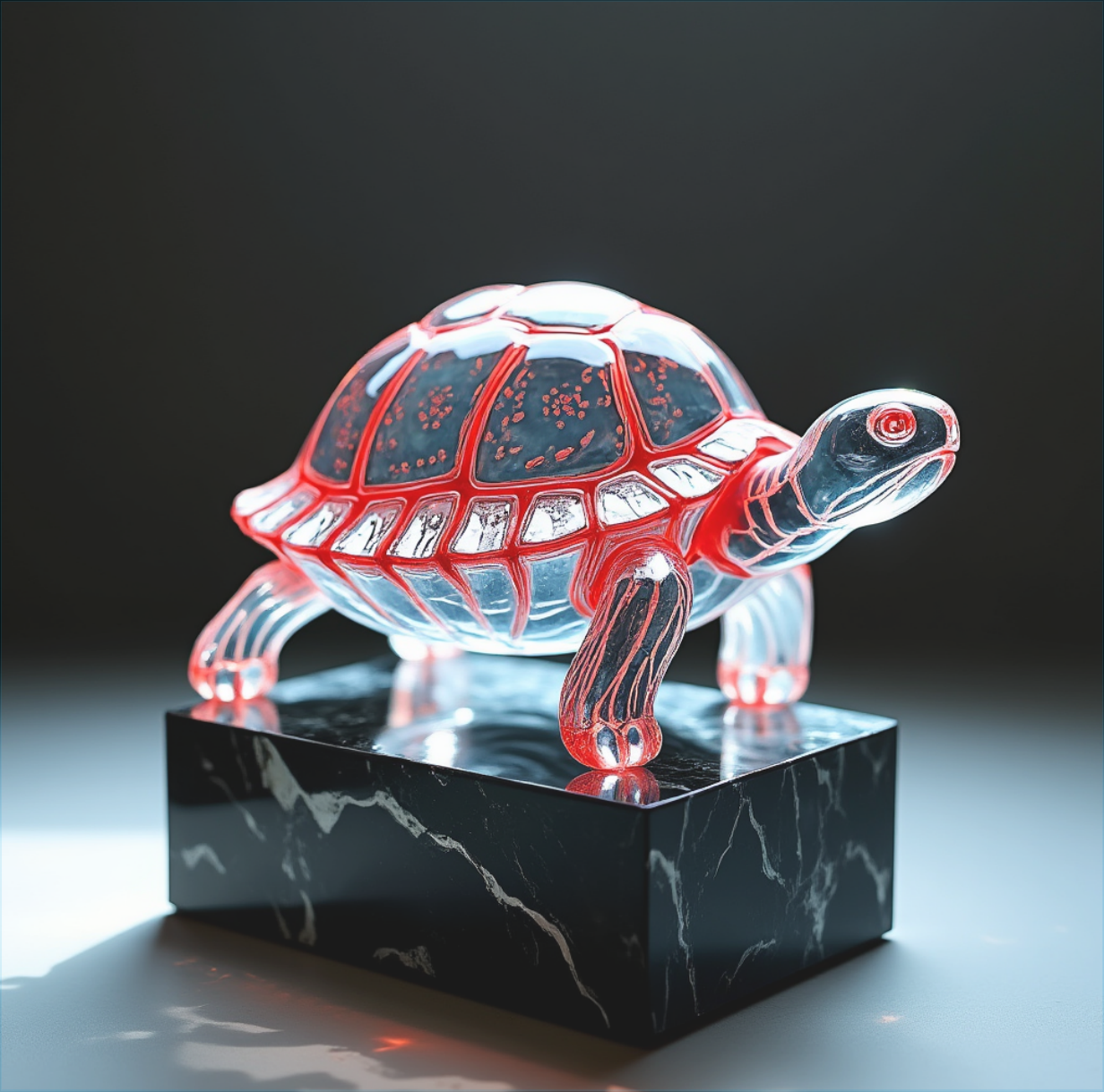} \\
        \midrule
        \addlinespace[5pt] % 增加行间距，防止拥挤

        % 第二行: Caption
        \textbf{Caption} & \textbf{Type1:} Infer the image-generation prompt from the picture. 
        \newline
        \textbf{Type2:}Please provide a caption for this picture.
        \newline
        \includegraphics[width=0.66\linewidth]{figures/figures_appendix/turtle_cropped.pdf}
        & A glass sculpture in the shape of a turtle with intricate patterns of red lines on its shell, resting on a black marble pedestal, with soft light coming from above, highlighting the contours of the turtle and casting delicate shadows on the floor. \\
        \midrule
        \addlinespace[5pt]

        % 第三行: Judgment
        \textbf{Judgement} & Judge the image quality based on the generation prompt. (See Fig.~\ref{fig:judger})       
        \newline
        \includegraphics[width=0.66\linewidth]{figures/figures_appendix/turtle_cropped.pdf}
        &
        \{  "analysis": "The generated image accurately depicts a glass sculpture in the shape of a turtle with intricate patterns of red and blue lines on its shell......(Lengthy analysis text omitted)",  "score": 10\}
         \\
        \midrule
        \addlinespace[5pt]

        % 第四行: Reflection
        \textbf{Reflection} & A glass sculpture in the shape of a turtle with intricate patterns of red lines on its shell, resting on a black marble pedestal, with soft light coming from above, highlighting the contours of the turtle and casting delicate shadows on the floor. &
        \includegraphics[width=0.4\linewidth]{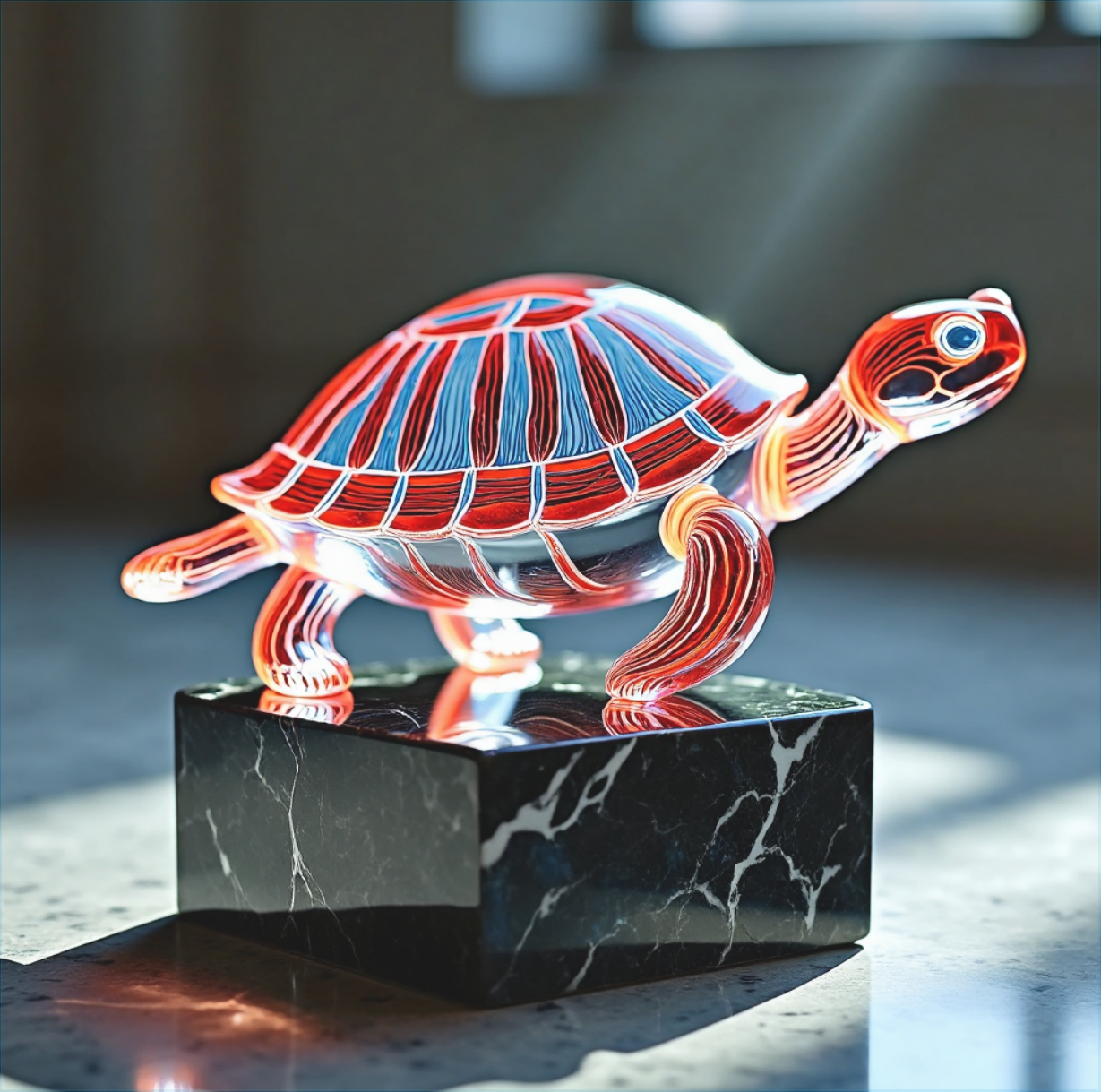}
        \hfill \raisebox{3em}{$\longrightarrow$} \hfill % 这是一个长箭头，左右自动填充间距
        \includegraphics[width=0.4\linewidth]{figures/figures_appendix/turtle_cropped.pdf} \\
        \bottomrule
    \end{tabularx}
    \caption{Examples of Generation, Caption, Judgement, Reflection training data. We choose the same image and prompt for better illustration.}
    \label{tab:train_data_example}
\end{table*}

%% file: tables/detail_tables.tex
\begin{table*}[ht]
    \centering
    \caption{\textbf{Quantitative evaluation results on OneIG-EN.} 
    }
    \resizebox{0.99\linewidth}{!}{
    \begin{tabular}{c|ccccc|c}
        \toprule
        \textbf{Model} & \textbf{Alignment}& \textbf{Text} & \textbf{Reasoning} & \textbf{Style}& \textbf{Diversity} & \textbf{Overall}$\uparrow$ \\
        \midrule
        Janus-Pro~\citep{januspro2025} & 0.553  & 0.001  &   0.139     & 0.276 & 0.365 & 0.267\\

        T2I-R1~\citep{jiang2025t2i} & 0.804 & 0.073 & 0.167 & 0.290 & 0.277 & 0.322\\
        BLIP3-o~\citep{chen2025blip3} & 0.711  & 0.013  &   0.223      & 0.361 & 0.229 & 0.307\\
        BAGEL~\citep{deng2025emerging} & 0.769  & 0.244  &   0.173    & 0.367 & 0.251& 0.361\\
        Show-o2-7B~\citep{xie2025show} & 0.817 & 0.002 & 0.226 & 0.317 & 0.177&0.308\\
        SDv1.5~\citep{rombach2022high} & 0.565 & 0.010 & 0.207 & 0.383 & 0.429 &0.319\\
        SDXL~\citep{sdxl} & 0.688 & 0.029 & 0.237 & 0.332 & 0.296 &0.316\\
        FLUX.1-dev~\citep{flux} & 0.786 & 0.523 & 0.253 & 0.368 & 0.238 & 0.434\\
        SD3~\citep{SD3} &0.805 & 0.407 & 0.293 & 0.386 & 0.244&0.427\\
        FLUX.1-dev~\citep{flux} & 0.786 & 0.523 & 0.253 & 0.368 & 0.238 & 0.434\\
        SANA-1.5 4.8B (PAG)~\citep{xie2025sana} & 0.765 & 0.069 & 0.217 & 0.401 & 0.216 &0.334\\
        Lumina-Image 2.0~\citep{qin2025lumina} & 0.819 & 0.106 & 0.270 & 0.354 & 0.216 & 0.353\\
        IRG* ~\citep{huang2025interleaving} 
        & 0.839 & 0.377 & 0.239 & 0.427 & 0.192 & 0.415 \\
        OmniGen2 ~\citep{xiao2025omnigen} 
        & 0.804 & 0.680 & 0.271 & 0.377 & 0.242 & 0.475 \\
  %       BAGEL-SRUM ~\citep{srum} 
  %       & 0.815  & 0.245 & 0.201 & 0.385 & 0.166 & 0.362 \\
  %       BAGEL-RecA ~\citep{xie2025reconstruction} 
  %       & 0.827 & 0.429 & 0.209 & 0.387 & 0.218 & 0.414
  % \\
        {\modelname}   & 
0.841 & 
0.468 & 
0.232 & 
0.395 & 
0.203&
0.426  \\
        \midrule
        \color[gray]{0.4}GPT-4o~\citep{openai2025chatgpt4o} & \color[gray]{0.4}0.851 & \color[gray]{0.4}0.857 & \color[gray]{0.4}0.345 & \color[gray]{0.4}0.462 & \color[gray]{0.4}0.151 & \color[gray]{0.4}0.533\\
        \bottomrule
    \end{tabular}
    }
    \label{tab:oneig_en}
\end{table*}

\begin{table*}[t]
\centering
\caption{\textbf{Quantitative evaluation results of instruct-following capability on TIIF testmini (QwenVL2.5-72B as the evaluation).} 
* indicates that the model has not yet been open-sourced; we report the metrics as presented in the official paper.
}
\renewcommand{\arraystretch}{1.7} 
\setlength{\tabcolsep}{3pt}

\centering
\resizebox{0.99\linewidth}{!}{
\begin{tabular}{c|cc|cc|cccccc|cc|cccccccccc|cc}
\toprule
\multirow{3}{*}{\textbf{Model}}
  & \multicolumn{2}{c|}{\multirow{2}{*}{\textbf{Overall}}}
  & \multicolumn{8}{c|}{\textbf{Basic Following}}
  & \multicolumn{12}{c|}{\textbf{Advanced Following}}
  & \multicolumn{2}{c}{\textbf{Designer}} \\

\cmidrule(lr){4-11} \cmidrule(lr){12-23} \cmidrule(lr){24-25}

& & &
  \multicolumn{2}{c|}{Avg}                    %
  & \multicolumn{2}{c}{Attribute}
  & \multicolumn{2}{c}{Relation}
  & \multicolumn{2}{c|}{Reasoning}
  & \multicolumn{2}{c|}{Avg}                  %
  & \multicolumn{2}{c}{\makecell{Attribute\\+Relation}}
  & \multicolumn{2}{c}{\makecell{Attribute\\+Reasoning}}
  & \multicolumn{2}{c}{\makecell{Relation\\+Reasoning}}
  & \multicolumn{2}{c}{Style}
  & \multicolumn{2}{c|}{Text}
  & \multicolumn{2}{c}{\makecell{Real\\World}} \\

& short & long &          %
  short & long &          %
  short & long &          %
  short & long &          %
  short & long &          %
  short & long &          %
  short & long &          %
  short & long &          %
  short & long &          %
  short & long &          %
  short & long &          %
  short & long            %
\\
\midrule

FLUX.1-dev~\citep{flux}  &66.24	&66.72	&74.41	&76.67	&72.50	&75.50	&78.20	&79.78	&72.52	&74.73	&60.72	&60.95	&66.76	&65.50	&61.76	&60.74	&56.60	&57.49	&63.33	&60.00	&44.49	&54.75	&74.63	&72.01 \\
FLUX.1-Pro~\citep{flux} &63.75	&63.53	&71.39	&73.57	&70.00	&68.50	&68.51	&79.97	&75.66	&72.23	&64.63	&61.42	&70.69	&72.99	&62.34	&57.27	&64.65	&57.11	&63.00	&63.00	&34.39	&36.65	&69.94	&66.78 \\
DALL-E 3~\citep{dalle3} &74.47	&72.94	&77.35	&78.40	&77.62	&75.00	&80.22	&79.67	&74.22	&80.54	&70.11	&68.45	&76.65	&75.05	&68.39	&68.07	&63.64	&59.92	&79.31	&80.00	&74.07	&75.51	&76.12	&62.69 \\
SD3.5-large~\citep{SD3} &68.69	&64.92	&73.72	&72.10	&77.50	&66.50	&74.79	&77.16	&68.87	&72.64	&65.59	&63.41	&70.85	&68.22	&65.03	&62.93	&61.03	&61.66	&56.67	&60.00	&73.30	&46.15	&70.15	&69.03 \\
PixArt-$\Sigma$~\citep{chen2024pixart} &57.46	&57.04	&67.74	&68.19	&65.50	&69.50	&74.33	&72.11	&63.40	&62.96	&56.71	&54.52	&62.47	&59.67	&57.51	&55.08	&54.84	&52.64	&76.67	&73.33	&2.71	&4.98	&63.06	&63.06 \\
Show-o~\citep{xie2024show} &57.34	&61.33	&69.99	&75.30	&66.50	&80.00	&76.47	&71.88	&67.00	&74.04	&58.25	&58.19	&67.21	&64.33	&54.26	&58.86	&61.38	&56.19	&46.67	&66.67	&4.98	&11.31	&71.64	&68.66 \\
Janus-Pro-7B~\citep{januspro2025} &65.38	&61.10	&74.99	&73.19	&74.50	&78.00	&73.69	&70.51	&76.77	&71.04	&61.77	&56.03	&65.71	&66.48	&62.01	&55.62	&61.16	&49.34	&43.33	&70.00	&38.46	&42.08	&79.48	&73.51 \\
T2I-R1~\citep{jiang2025t2i} &67.61	&68.34	&81.14	&79.45	&80.50	&78.50	&83.09	&79.49	&79.81	&80.37	&67.38	&65.90	&69.92	&65.27	&70.10	&71.62	&68.69	&64.68	&50.00	&63.33	&32.13	&37.56	&74.25	&74.25 \\
BAGEL~\citep{deng2025emerging}&70.97  &71.79  &78.16  &78.12  &78.00  &79.50  &80.24  &79.08  &76.25  &75.77  &68.23  &68.19  &73.37  &77.49  &64.36  &66.15  &68.92  &61.48  &80.00  &80.00  &40.72  &52.40  &76.87  &74.63  \\
MidJourney v7~\citep{midjourneyv7} &65.92	&62.43	&73.96	&74.63	&75.00	&82.00	&78.74	&78.51	&68.12	&68.55	&63.44	&62.59	&70.60	&74.03	&64.43	&59.58	&58.84	&61.34	&66.67	&33.33	&31.67	&34.39	&79.22	&75.32 \\
Show-o2 ~\citep{xie2025show} &62.80	&63.87	&75.30	&74.45	&73.00	&71.00	&77.22	&74.09	&75.69  &78.25	&61.38 	&66.12	&63.47  	&67.44	&62.63  &70.31	&64.15 	&60.00  &60.00	&33.33	&14.03  	&10.86	&75.00	&74.63 \\
BAGEL~\citep{deng2025emerging}&68.06 &68.78  &77.63  &79.40  &75.00  &77.00  &78.55  &82.37  &79.33  &78.81  &71.24  &68.20  &77.65  &75.37  &69.77  &65.87  &72.93  &67.91  &69.93  &63.33  &26.24  &26.70   &69.78  &71.64  \\
IRG* ~\citep{huang2025interleaving}  & 76.00 & 73.77 & 83.17 & 81.28 & 81.00 & 76.00 & 82.96 & 81.86 & 85.54 & 85.98 & 75.25 & 74.66 & 75.82 & 77.25 & 78.16 & 77.76 & 73.84 & 72.93 & 90.00 & 70.00 & 43.89 & 47.51 & 72.76 & 74.63 \\

{\modelname}&
 74.70&72.94	&79.43	&78.53	&81.50	&	79.50&83.14	&79.84	&73.64	&76.25  &73.39	&71.81 	&76.84	&74.59  	&72.34	&71.33  &72.81	&71.66 	& 73.33 &76.67	& 58.85  &49.77  	&79.85	&	76.87  \\
\midrule
\color[gray]{0.4}GPT-4o~\citep{openai2025chatgpt4o} &\color[gray]{0.4}84.19	&\color[gray]{0.4}84.61	&\color[gray]{0.4}85.30	&\color[gray]{0.4}86.55	&\color[gray]{0.4}81.00	&\color[gray]{0.4}82.12	&\color[gray]{0.4}86.16	&\color[gray]{0.4}84.12	&\color[gray]{0.4}88.74	&\color[gray]{0.4}94.50	&\color[gray]{0.4}81.24	&\color[gray]{0.4}79.75	&\color[gray]{0.4}81.95	&\color[gray]{0.4}81.55	&\color[gray]{0.4}80.03	&\color[gray]{0.4}79.85	&\color[gray]{0.4}80.88	&\color[gray]{0.4}75.68	&\color[gray]{0.4}76.67	&\color[gray]{0.4}86.67	&\color[gray]{0.4}92.76	&\color[gray]{0.4}90.05	&\color[gray]{0.4}89.55	&\color[gray]{0.4}88.06  \\
\bottomrule
\end{tabular}
}
\label{tab:tiif_qwen}
\end{table*}
\begin{table*}[t]
    \centering
    \scriptsize
    \caption{\textbf{Comparison of world knowledge reasoning on WISE.} WISE examines the complex semantic understanding and world knowledge for T2I generation. `Gen. Only' stands for an image generation model, and `Unified' denotes a model that has both understanding and generation capabilities.
    * indicates that the model has not yet been open-sourced; we report the metrics as presented in the official paper.
    }
    \label{tab:wisescore}
    \resizebox{0.99\linewidth}{!}{
    \begin{tabular}{c|c|cccccc|c}
    \toprule
    \textbf{Type} & \textbf{Model} & \textbf{Cultural}  & \textbf{Time}     & \textbf{Space}    & \textbf{Biology}    & \textbf{Physics} & \textbf{Chemistry} & \textbf{Overall$\uparrow$} \\
    \midrule
            \multirow{6}{*}{\rotatebox{90}{\textit{Gen. Only}}} &
 SDv1.5~\citep{rombach2022high} & 0.34 & 0.35& 0.32&0.28 &0.29 &0.21 & 0.32\\
& SDXL~\citep{sdxl} &0.43  & 0.48 &0.47  &0.44  &0.45 &0.27 & 0.43 \\
& SD3.5-large~\citep{SD3} & 0.44 &0.50 &0.58  & 0.44&0.52 &0.31 & 0.46 \\
& PixArt-Alpha~\citep{chen2024pixart} & 0.45  & 0.50& 0.48 & 0.49& 0.56 &0.34 & 0.47\\
& playground-v2.5~\citep{li2024playground} & 0.49  &0.58  & 0.55&0.43  & 0.48&0.33 & 0.49 \\
& FLUX.1-dev~\citep{flux} & 0.48  &0.58 &0.62  &0.42  &0.51 & 0.35 & 0.50 \\
    \midrule
    \multirow{10}{*}{\rotatebox{90}{\textit{Unified}}} 
& Janus~\citep{wu2025janus} &0.16 &0.26 &0.35 & 0.28 &0.30 & 0.14& 0.23\\
& Show-o-512~\citep{xie2024show} & 0.28 &0.40  &0.48 & 0.30& 0.46 & 0.30 & 0.35\\
& Janus-Pro-7B~\citep{januspro2025} & 0.30& 0.37& 0.49 & 0.36&0.42 &0.26 & 0.35 \\
& Emu3~\citep{emu3} & 0.34 & 0.45 & 0.48 & 0.41  & 0.45 & 0.27 & 0.39 \\
& MetaQuery-XL~\citep{pan2025transfer} & 0.56& 0.55 &0.62 &  0.49 &  0.63 & 0.41 & 0.55 \\
& BAGEL~\citep{deng2025emerging} & 0.42 & 0.53 & 0.64 & 0.42 & 0.57 & 0.43 & 0.50 \\
& Show-o2~\citep{xie2025show} & 0.64 & 0.58 & 0.61 & 0.58 & 0.63 & 0.49 & 0.61 \\
& T2I-R1 ~\citep{jiang2025t2i}  & 0.56 & 0.55 & 0.63 & 0.54 & 0.55 & 0.30 & 0.54 \\
& BLIP3-o ~\citep{blip3o}  & 0.49 & 0.51 & 0.63 & 0.54 & 0.63 & 0.37 & 0.52 \\

& {\modelname} &0.48	&0.56&	0.67&	0.47&	0.67	&0.47& 0.55 \\
\midrule
& \color[gray]{0.4}GPT-4o~\citep{openai2025chatgpt4o} & \color[gray]{0.4}0.81 & \color[gray]{0.4}0.71 & \color[gray]{0.4}0.89 & \color[gray]{0.4}0.83 & \color[gray]{0.4}0.79 & \color[gray]{0.4}0.74 & \color[gray]{0.4}0.80 \\
        \bottomrule
    \end{tabular}
    }
\end{table*}
\begin{table*}[ht]
    \centering
    \setlength{\tabcolsep}{2pt}
    \renewcommand{\arraystretch}{1.3}
    \small
    \caption{\textbf{Comprehensive T2I-CompBench Results.} This table includes T2I~\citep{flux,SD3,sdxl} and UMMs~\citep{januspro2025,xie2025show}.}
            \vspace{-0.5em}
    \label{tab:compbench_final}
\begin{tabular}{lccccccccc}
    \toprule
    \textbf{Model} & \textbf{3d Spatial} & \textbf{Color} & \textbf{Complex} & \textbf{Nonspatial} & \textbf{Numeracy} & \textbf{Shape} & \textbf{Spatial} & \textbf{Texture} & \textbf{Overall} \\
    \midrule
    \multicolumn{10}{c}{\textit{T2I Models}} \\ \midrule
    FLUX.1-dev & 76.39 & 90.63 & 83.51 &\bfseries 87.47 & \bfseries75.30 & 80.20 & 84.23 & 87.07 & 83.10 \\
    FLUX.1-schnell & \bfseries 79.38 & 84.53 & 81.96 & 85.55 & 72.82 & 82.20 & 85.49 & 86.38 & 82.29 \\
    SD-3-medium & 77.83 & \bfseries91.63 & \bfseries84.73 & 86.12 & 72.80 & \bfseries83.72 & \bfseries 88.20 & \bfseries89.03 & \bfseries84.26 \\
    SD-xl-base-1 & 72.25 & 77.75 & 75.00 & 85.28 & 57.14 & 72.18 & 77.08 & 78.38 & 74.38 \\
    \midrule \multicolumn{10}{c}{\textit{Unified Multimodal Models}} \\ \midrule
    Janus-Pro & 76.17 & 84.25 & 80.28 & 80.47 & 56.43 & 65.14 & 79.67 & 69.67 & 74.01 \\
    T2I-R1 &79.35 & 92.11 & 85.48 & 83.32 & 69.47 & 74.08 & 86.44 & 84.85 & 81.89\\

    Show-O2 & \bfseries 88.61 & 87.73 & 87.88 & 85.91 & 69.74 & 73.99 & 86.60 & 82.17 & 82.83 \\
    OmniGen2 & 82.21 & 92.22 & 86.87 & 88.51 & 72.00 & 83.95 & 90.07 & \bfseries 90.88 & 85.84 \\

    BLIP3o & 81.73 & 89.92 & 85.55 & 84.78 & 71.67 & 83.75 & 92.47 & 87.45 & 84.66 \\
    BAGEL & 77.98 & 89.30 & 83.32 & 85.03 & 70.40 & 81.94 & 81.52 & 87.93 & 82.18 \\

    {\modelname} &84.12&93.92&88.80&89.50&83.47&87.07&88.92&91.48& 88.51 \\
  
    \bottomrule
\end{tabular}
\end{table*}

\begin{table*}[t]
    \centering
    \scriptsize
    \caption{\textbf{Evaluation of text-to-image generation ability on GenEval benchmark.} `Gen. Only' stands for an image generation model, and `Unified' denotes a model that has both understanding and generation capabilities.
    $\dagger$ refer to the methods using MLLM rewriter.The best Overall results are \textbf{bolded}.
    }
    \resizebox{0.99\linewidth}{!}{
    \begin{tabular}{c|c|cccccc|c}
        \toprule
        \textbf{Type} & \textbf{Model}  & \textbf{Single Obj.} & \textbf{Two Obj.} & \textbf{Counting} & \textbf{Colors} & \textbf{Position} & \textbf{Color Attri.} & \textbf{Overall$\uparrow$} \\
        \midrule
        \multirow{8}{*}{\rotatebox{90}{\textit{Gen. Only}}}
        & PixArt-$\alpha$~\citep{chen2024pixart} &  0.98 & 0.50 & 0.44 & 0.80 & 0.08 & 0.07 & 0.48 \\
        & SDv$2.1$~\citep{rombach2022high} & 0.98 & 0.51 & 0.44 & 0.85 & 0.07 & 0.17 & 0.50 \\
        & DALL-E $2$~\citep{dalle2}  & 0.94 & 0.66 & 0.49 & 0.77 & 0.10 & 0.19 & 0.52 \\
        & Emu$3$-Gen ~\citep{emu3}  & 0.98 & 0.71 & 0.34 & 0.81 & 0.17 & 0.21 & 0.54 \\
        & SDXL~\citep{sdxl} &  0.98 & 0.74 & 0.39 & 0.85 & 0.15 & 0.23 & 0.55 \\
        & DALL-E $3$~\citep{dalle3} & 0.96 & 0.87 & 0.47 & 0.83 & 0.43 & 0.45 & 0.67 \\
        & SD3-Medium~\citep{SD3} & 0.99 & 0.94 & 0.72 & 0.89 & 0.33 & 0.60 & 0.74 \\
        & FLUX.1-dev$^{\dagger}$~\citep{flux} & 0.98 & 0.93 & 0.75 & 0.93 & 0.68 & 0.65 & 0.82 \\
        \midrule
        \multirow{17}{*}{\rotatebox{90}{\textit{Unified}}}
        & Chameleon~\citep{chameleon} &  - & - & - & - & - & - & 0.39 \\
        & LWM~\citep{lwm} &  0.93 & 0.41 & 0.46 & 0.79 & 0.09 & 0.15 & 0.47 \\
        & SEED-X~\citep{seed-x}  & 0.97 & 0.58 & 0.26 & 0.80 & 0.19 & 0.14 & 0.49 \\
        & TokenFlow-XL~\citep{qu2024tokenflow} &  0.95 & 0.60 & 0.41 & 0.81 & 0.16 & 0.24 & 0.55 \\
        & ILLUME~\citep{wang2024illume} &  0.99 & 0.86 & 0.45 & 0.71 & 0.39 & 0.28 & 0.61 \\
        & Janus~\citep{wu2025janus} & 0.97 & 0.68 & 0.30 & 0.84 & 0.46 & 0.42 & 0.61 \\
        & Transfusion~\citep{transfusion} & - & - & - & - & - & - & 0.63 \\
        & Emu$3$-Gen$^{\dagger}$\citep{emu3} & 0.99 & 0.81 & 0.42 & 0.80 & 0.49 & 0.45 & 0.66 \\
        & Show-o~\citep{xie2024show} &  0.98 & 0.80 & 0.66 & 0.84 & 0.31 & 0.50 & 0.68 \\
        & Janus-Pro-7B~\citep{januspro2025} &  0.99 & 0.89 & 0.59 & 0.90 & 0.79 & 0.66 & 0.80 \\
        & MetaQuery-XL$^{\dagger}$~\citep{pan2025transfer} &  -& - & - & -& -& -& 0.80 \\
        & BAGEL~\citep{deng2025emerging} & 0.99 & 0.95  & 0.76 & 0.87 & 0.50 & 0.60 & 0.78 \\
        & Show-o2~\citep{xie2025show} &  1.00 & 0.87 & 0.58 & 0.92 & 0.52 & 0.62 & 0.76 \\
    & BAGEL~\citep{deng2025emerging} & 0.99 & 0.92  & 0.75 & 0.89 & 0.54 & 0.63 & 0.79 \\
    & IRG* ~\citep{huang2025interleaving} & 0.98 & 0.94  & 0.83 & 0.86 & 0.74 & 0.73 & 0.85 \\
    & UniGen* ~\citep{tian2025unigen}   & 1.00 & 0.94 & 0.78 & 0.87 & 0.57 & 0.54 & 0.78 \\
    & UniRL ~\citep{unirl}   & 0.96 & 0.80 & 0.67 & 0.86 & 0.50 & 0.67 & 0.74 \\

        & {\modelname} & 0.99  & 0.94 & 0.80 & 0.88 & 0.61 & 0.73 & 0.82 \\
    \midrule
    & \color[gray]{0.4}GPT-4o~\citep{openai2025chatgpt4o} & \color[gray]{0.4}0.99 & \color[gray]{0.4}0.92 & \color[gray]{0.4}0.85 & \color[gray]{0.4}0.92 & \color[gray]{0.4}0.75 & \color[gray]{0.4}0.61 & \color[gray]{0.4}0.84 \\
    \bottomrule
    \end{tabular}
    }
    \label{tab:geneval}
\end{table*}

\begin{table*}[t]
    \centering
    \scriptsize
    \caption{Quantitative evaluation results on DPG}
    \small
    \begin{tabular}{l|ccccc|c}
    \toprule
    \textbf{Model}           & \bf Global & \bf Entity & \bf Attribute & \bf Relation & \bf Other & \bf Overall$\uparrow$ \\
    \midrule
    PixArt-$\alpha$~\citep{chen2024pixart}  & 74.97  & 79.32  & 78.60      & 82.57    & 76.96 & 71.11    \\
    Lumina-Next~\citep{zhuo2024luminanext}  & 82.82  & 88.65  & 86.44     & 80.53    & 81.82 & 74.63    \\
    Playground v2.5~\citep{li2024playground}  & 83.06  & 82.59  & 81.20      & 84.08    & 83.50  & 75.47    \\
    Hunyuan-DiT~\citep{li2024hunyuandit}      & 84.59  & 80.59  & 88.01     & 74.36    & 86.41 & 78.87    \\
    Janus~\citep{wu2025janus}     & 82.33  & 87.38  & 87.70      & 85.46    & 86.41 & 79.68    \\
    Janus-Pro-1B~\citep{chen2025janus}     & 87.58  & 88.63  & 88.17     & 88.98    & 88.30  & 82.63    \\
    DALL-E 3~\citep{dalle3}     & 90.97  & 89.61  & 88.39     & 90.58    & 89.83 & 83.50     \\
    FLUX.1-dev ~\citep{flux}    & 74.35  & 90.00     & 88.96     & 90.87    & 88.33 & 83.84    \\
    SD3 Medium~\citep{SD3}    & 87.90   & 91.01  & 88.83     & 80.70   & 88.68 & 84.08    \\
    Janus-Pro-7B~\citep{chen2025janus}     & 86.90   & 88.90   & 89.40    & 89.32    & 89.48 & 84.19    \\
    BAGEL  ~\citep{deng2025emerging} & -  & - & - & - & - & 84.03 \\

    {\modelname}   &  91.62 & 91.97 & 91.39 & 91.22 & 91.64 & 86.83 \\
    \bottomrule
    \end{tabular}\label{tab:dpg}
\end{table*}

%% file: tables/generation_prompt.tex
\begin{figure*}[!ht] 
\begin{AIbox}{Prompt for Proposer}
{\color{black}\bf System Prompt:} \\
\textbf{Character Introduction}  \\
You are a specialist dataset architect for PromptBench. Your mission is to synthesize high-quality, high-complexity text-to-image prompts that push the limits of generative models.

\textbf{Your Task}  \\
-\textbf{Target Category}: \\
Generate prompt \textbf{ONLY} for the category defined by: \{\textit{major category}\}.\\
-\textbf{Category Definition and Specific Rule}(MUST FOLLOW THE RULE FOR THE TARGET CATEGORY): \\
\{\textit{category rule}\}

-\textbf{Informational Density}: \\
The prompt \textbf{must contain sufficient descriptive detail} to ensure complex image generation. Do not prioritize brevity over informational density\\
% -\textbf{Conceptual Challenge}: Avoid generic descriptions. Aim for nuanced, multi-layered scenarios that require high visual reasoning from the T2I model.\\
% -\textbf{Format Integrity}: \\
% The output must be a \textbf{strictly valid JSON list}.\\
\textbf{Response Format} \\
\textbf{Strictly follow the JSON format} to output only the modified dialog without redundancy, and do not add comments (//) in the response. \\
\begin{lstlisting}[style=prompt]
{
"major_category": "The primary classification", 
"subcategory": "The secondary classification"
"prompt": "The high-density descriptive instruction."
}
\end{lstlisting}

\textbf{Example}  \\
\{\textit{Few-shot Example}\} \\

{\color{black}\bf User Prompt:} \\
Generate exactly ONE new prompt. \\
Target Major Category: \{\textit{major category}\}. \\
Target Subcategory: \{\textit{subcategory}\} \\
Each generated item must have a \textbf{major\_category} field set to \{\textit{major category}\}, a \textbf{subcategory} field set to \{\textit{subcategory}\}, and a \textbf{prompt} field. Ensure high diversity and strictly adhere to the rule.

\end{AIbox} 
\vspace{-1em}
\caption{The prompt template for prompt proposer.}
\label{fig: proposer}
\vspace{-1em}
\end{figure*}

%% file: tables/judge_prompt.tex
\begin{figure*}[!ht] 
\begin{AIbox}{Prompt for Image Judge}
{\color{black}\bf System Role:} \\
You are a rigorous \textbf{Visual Quality Assessment Expert}. Your mission is to evaluate the alignment and technical fidelity of generated images against specific text prompts using a deterministic, objective framework.

\textbf{Evaluation Criteria (Ranked by Priority):} \\
% \begin{enumerate}
%     \item \textbf{Instruction Adherence}: Does the image faithfully execute the theme, quantity, key attributes, and style? Verify the inclusion/exclusion of all specified elements.
%     \item \textbf{Logical \& Physical Integrity}: Does the image conform to reality? Check for anatomical accuracy, correct perspective, physical consistency, and legible/correctly spelled text.
%     \item \textbf{Technical Execution}: Assess clarity, noise, detail (especially hands/faces), lighting, and composition. Ensure no watermarks or improper cropping.
% \end{enumerate}
\{\textit{category specific Judgement Rubrics }\} \\

\textbf{Scoring Standard:(0 - 10)}  \\
% \begin{itemize}
%     \item \textbf{0--3}: Severe instruction deviation or fatal logical/anatomical errors.
%     \item \textbf{4--6}: Basic compliance but suffers from multiple notable issues.
%     \item \textbf{7--8}: High compliance with only minor technical or stylistic artifacts.
%     \item \textbf{9--10}: Near-perfect or perfect execution with exceptional detail and fidelity.
% \end{itemize}
\{\textit{category specific scoring standard}\}

\textbf{Response Format:} \\
Return a \textbf{strictly valid JSON object} only. Do not include conversational filler, markdown commentary, or code block delimiters.
\begin{lstlisting}[style=prompt]
{
  "analysis": "A concise, objective breakdown of the evaluation points.",
  "score": "Integer or float from 0 to 10"
}
\end{lstlisting}

\textbf{Input Data:} \\
Category: \{\textit{major category}\} \\
Prompt: \{\textit{prompt}\}\\
Image: [\textit{Image}]
\end{AIbox} 
\vspace{-1em}
\caption{The prompt template for reward judger.}
\label{fig:judger}
\vspace{-1em}
\end{figure*}

%% file: tables/compare_self.tex
\begin{table*}[t]
\small
\renewcommand{\arraystretch}{1.2}  % 缩小行高
\setlength{\tabcolsep}{3pt} 
\centering
\begin{tabular}{lcccc}
\hline
\textbf{Method} 
& \textbf{External Model Free} 
& \textbf{External Data Free} 
& \textbf{External Model} 
& \textbf{Hyperparameters↓}\\
\hline
IRG 
&  \xmark 
&  \xmark  
&   GPT-4o+Qwen2.5VL
&  0 \\
UniRL 
& \cmark
& \cmark 
& GPT-4o 
& 1\\

SRUM 
&  \xmark 
& \xmark 
& SAM3 
& 1\\

RecA 
&  \cmark 
& \xmark 
& GPT-4o
& 3\\

{\modelname}
& \cmark
& \cmark 
& -
& 0\\
\hline
\end{tabular}
\caption{Comparison of different methods in terms of external dependencies and prompt construction strategies. \textbf{Without relying on external task-specific models or annotated data, {\modelname} achieves state-of-the-art performance on OneIG-EN using only 5K training samples.}}
\label{tab: model_compare}
\end{table*}

%% file: tables/eval_bench_prompt.tex
\begin{figure*}[!t]
\centering
\begin{minipage}{0.95\textwidth}
\begin{AIbox}{Prompt for \textbf{UniCycle} evaluation (non-text task type)}
\small
You are a strict visual QA evaluation assistant.

\textbf{You will be given:}

1) TASK\_TYPE: the evaluation dimension to consider.

2) IMAGE\_PROMPT describing what the image should contain.

3) ONE QA pair (Question, Answer).

4) A Reference Answer.

\textbf{Your Task}\\
Determine whether the Answer is consistent with IMAGE\_PROMPT for TASK\_TYPE only.

Ignore all other aspects. You may use the Reference Answer only for equivalence checking.

\textbf{Rules}\\
- Use ONLY IMAGE\_PROMPT; do NOT use external knowledge.\\
- Output "yes" ONLY if IMAGE\_PROMPT clearly supports the Answer for TASK\_TYPE.\\
- Output "no" if the Answer contradicts IMAGE\_PROMPT, or if IMAGE\_PROMPT is insufficient.\\
- Output "no" if the Answer is a refusal, uncertainty, or hedging.\\
- Be strict: required details must be explicitly supported.\\
- Do NOT explain. Output JSON only.

\textbf{Normalization rules (for equivalence checking only)}\\
- Ignore letter case, punctuation, and extra whitespace.\\
- Minor spelling variants are equivalent (e.g., gray/grey, color/colour).

Output JSON with exactly these keys:

\{
  "question": "<question>",\\
  "answer": "<answer>",\\
  "evaluation": "yes" or "no" \\
\}

[TASK\_TYPE]\\
\{task\_type\}

[IMAGE\_PROMPT]\\
\{image\_prompt\}

Question: \{question\}\\
Answer: \{answer\}\\
Reference Answer: \{refer\_ans\}
\end{AIbox}
\end{minipage}
\caption{The prompt template for {\evalname} evaluation(non-text task type).}
\label{fig:evaluation}
\end{figure*}

%% file: tables/eval_bench_prompt_text.tex
\begin{figure*}[t]
\centering
\small

% ================== (1) AIbox ==================
\begin{minipage}{0.95\textwidth}
\begin{AIbox}{Prompt for \textbf{UniCycle} evaluation (text task type)}
\small
You are a strict text rendering QA evaluator.

\textbf{You will be given:}\\
1) IMAGE\_PROMPT describing what the image should contain\\
2) ONE QA pair (Question, Answer)\\
3) A Reference Answer

\textbf{Your task:}\\
Count how many required words in the Answer are correctly supported by
IMAGE\_PROMPT and Reference Answer.

Use ONLY IMAGE\_PROMPT. Do NOT use external knowledge.

Output JSON only with exactly these keys:
\{
  "question": "<question>",
  "answer": "<answer>",
  "evaluation": "<number of correctly answered words>"
\}

[IMAGE\_PROMPT]\\
\{image\_prompt\}

Question: \{question\}\\
Answer: \{answer\}\\
Reference Answer: \{refer\_ans\}
\end{AIbox}
\end{minipage}

\vspace{0.8em} % AIbox -> caption

\caption{The prompt template for {\evalname} evaluation (text task type).}
\label{fig:evaluation_text}

\vspace{4.2em} % Figure -> Table section break

% ================== (2) Table 1 ==================
\renewcommand{\arraystretch}{0.95}
\setlength{\tabcolsep}{4pt}
\begin{minipage}{0.9\textwidth}
\centering
\begin{tabular}{lcc}
\hline
Question Type & Count & Ratio (\%) \\
\hline
Total questions   & 2968 & 100.00 \\
MCQ questions     & 1067 & 35.95 \\
Yes/No questions  & 200  & 6.74  \\
Open-ended questions   & 1701 & 57.31 \\
\hline
\end{tabular}
\captionof{table}{Question types distribution of \evalname{}.}
\label{tab:question-distribution}
\end{minipage}

\vspace{4.2em} % Table 1 -> Table 2

% ================== (3) Table 2 ==================
\begin{minipage}{0.9\textwidth}
\centering
\begin{tabular}{lcccccc}
\hline
Model & Bagel & Show-o2 & Janus-Pro & UniCorn* & {\modelname} \\
\hline
Soft score & 58.2 & 52.5 & 25.8 & 58.6 & \textbf{66.6} \\
\hline
\end{tabular}
\captionof{table}{\textbf{Soft score} results on \evalname{}.}
\label{tab:soft-scores}
\end{minipage}

\end{figure*}